%% file: Manuscript.tex
\documentclass[lettersize,journal]{IEEEtran}
\usepackage{amsmath,amsfonts}
\usepackage{algorithmic}
\usepackage{algorithm}
\usepackage{array}
\usepackage{tabu,multirow}
\usepackage[table,dvipsnames]{xcolor}
\usepackage{color,colortbl}
\definecolor{Gray}{gray}{0.9}
\usepackage[caption=false,font=normalsize,labelfont=sf,textfont=sf]{subfig}
\usepackage{booktabs,makecell}
\usepackage{graphicx}
\usepackage{textcomp}
\usepackage{stfloats}
\usepackage{url}
\usepackage{verbatim}
\usepackage{graphicx}
\usepackage{cite}
\usepackage{caption}
\usepackage{bbm}
\makeatletter
\def\ps@IEEEtitlepagestyle{
  \def\@oddfoot{\mycopyrightnotice}
  \def\@evenfoot{}
}
\def\mycopyrightnotice{
  {\footnotesize
  \begin{minipage}{\textwidth}
  \centering
  Copyright~\copyright~2024 IEEE. Personal use of this material is permitted. However, permission to use this material for any other purposes must be obtained from the IEEE by sending an email to pubs-permissions@ieee.org.
  \end{minipage}
  }
}

\newcommand{\refFig}[1]{Fig. \ref{#1}}
\newcommand{\refTab}[1]{Table \ref{#1}}
\begin{document}
   
\title{Fine-grained Background Representation for Weakly Supervised Semantic Segmentation}

\author{Xu Yin, Woobin Im, Dongbo Min, Yuchi Huo, Fei Pan, Sung-Eui Yoon
\thanks{This work was supported by the National Research Foundation of Korea(NRF) grant funded by the Korea government(MSIT) (No. RS-2023-00208506(2024)) and was partially supported by the National Key R\&D Program of China (No. 2024YDLN0011) and NSFC (No. 62441205) (Corresponding authors: Sung-eui Yoon.)

Xu Yin and Woobin Im are with the School of Computing, Korea Advanced Institute of Science and Technology, Daejeon 34141, South Korea (E-mail: yinofsgvr@kaist.ac.kr and iwbn@kaist.ac.kr).

Dongbo Min is the Faculty of the Department of Computer Science and Engineering, Ewha Womans University, Seoul 03760, South Korea (E-mail: dbmin@ewha.ac.kr).

Yuchi Huo is with the State Key Lab of CAD and CG, Zhejiang University, China and Zhejiang Lab, China 310058 (E-mail: huo.yuchi.sc@gmail.com).

Fei Pan is with the School of Computer Science Engineering, University of Michigan. Email: feipan@umich.edu).

Sung-eui Yoon is with the Faculty of School of Computing, Korea Advanced Institute of Science and Technology, Deajeon 34141, South Korea (E-mail:
sungeui@gmail.com).
}}

\maketitle

\input{Content/Abstract}

\begin{IEEEkeywords}
Contrastive learning, Fine-grained background representation, Weakly supervised image segmentation.
\end{IEEEkeywords}
\section{Introduction}
\label{sec:introduction}
\input{Content/Introduction}

\section{Related Work}
\input{Content/Relatedwork_section}

\section{Methodology}

\input{Content/Method_section}
\section{Experiments}

\input{Content/Experiment_section}

\section{Conclusion}
\input{Content/Conclusion}

\appendix
\input{Content/Appendix}

\bibliography{Manuscript}
\bibliographystyle{IEEEtranS}

\end{document}

%% file: Content/Abstract.tex
\begin{abstract}
Generating reliable pseudo masks from image-level labels is challenging in the weakly supervised semantic segmentation (WSSS) task due to the lack of spatial information. Prevalent class activation map (CAM)-based solutions are challenged to discriminate the foreground (FG) objects from the suspicious background (BG) pixels (a.k.a. co-occurring) and learn the integral object regions. This paper proposes a simple fine-grained background representation (\textbf{FBR}) method to discover and represent diverse BG semantics and address the co-occurring problems. We abandon using the class prototype or pixel-level features for BG representation. Instead, we develop a novel primitive, negative region of interest (\textbf{NROI}), to capture the fine-grained BG semantic information and conduct the pixel-to-NROI contrast to distinguish the confusing BG pixels. We also present an active sampling strategy to mine the FG negatives on-the-fly, enabling efficient pixel-to-pixel intra-foreground contrastive learning to activate the entire object region. Thanks to the simplicity of design and convenience in use, our proposed method can be seamlessly plugged into various models, yielding new state-of-the-art results under various WSSS settings across benchmarks. Leveraging solely image-level (I) labels as supervision, our method achieves 73.2 mIoU and 45.6 mIoU segmentation results on Pascal Voc and MS COCO test sets, respectively. Furthermore, by incorporating saliency maps as an additional supervision signal (I+S), we attain 74.9 mIoU on Pascal Voc test set. Concurrently, our FBR approach demonstrates meaningful performance gains in weakly-supervised instance segmentation (WSIS) tasks, showcasing its robustness and strong generalization capabilities across diverse domains.
\end{abstract}

%% file: Content/Introduction.tex
\IEEEPARstart{F}{ully}-supervised semantic segmentation~\cite{EPL,tcsvt_3} requires a pixel-annotated training set, which is costly and time-consuming to create. Thus, great efforts have been put into weakly supervised semantic segmentation to reduce the annotation cost by leveraging less expensive yet weakly spatially-informative supervision signals, such as image tags~\cite{cian}, bounding boxes~\cite{boxinst}, and scribbles~\cite{universal_wsss,review2}. Current studies~\cite{AMN, advcam, eps} usually start from generating class activation maps~\cite{sipe} by training a classification network to build the seeds and then utilize refinement techniques~\cite{psa,irn}, for generating reliable pseudo masks, which are finally used to train the segmentation model~\cite{tcsvt_1,tcsvt_2}. In this work, we concentrate on the image-level weakly supervised semantic segmentation, where only the images’ class labels are given; we aim to enhance the quality of seeds and, thus, the segmentation results.
\begin{figure}[t]
\begin{center}
\includegraphics[width=8.3cm]{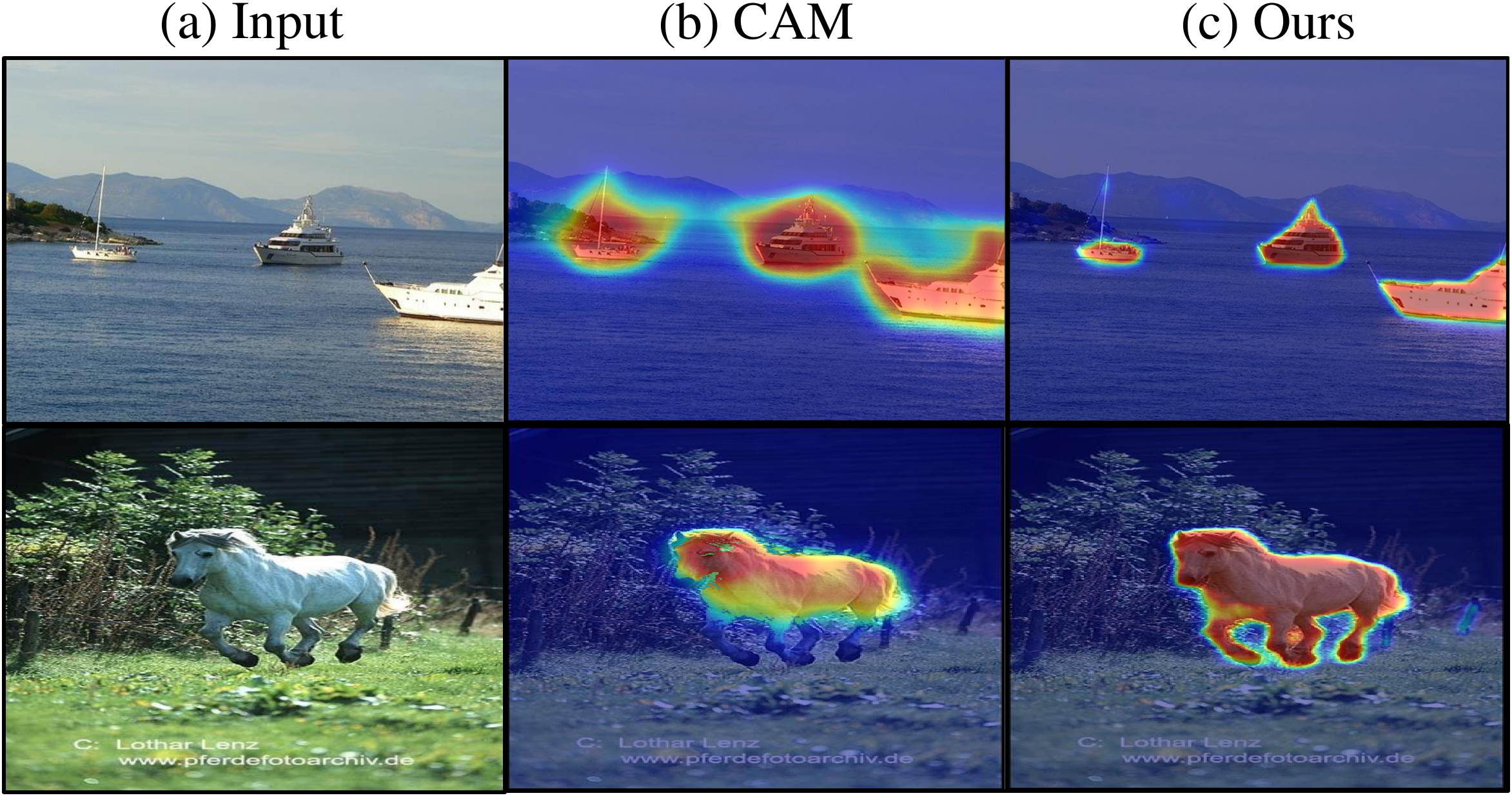}
\end{center}
 \vspace{-10pt}
   \caption{(a) Input (b) Class activation maps
via AMN~\cite{AMN} (c) Refined class activation maps with our method (on AMN). In
the 1st row (b), class activation maps mistake the lake (co-occurring background semantic) as the boat; in the 2nd row (b),
the horse is not completely activated.  
 }
   \label{fig:problem_definition}
   \vspace{-15pt}
\end{figure}

First, we illustrate two common problems of class activation maps (shown in \refFig{fig:problem_definition}): co-occurring background semantics and incomplete object region. The former refers to the disturbance from the confusing image background information for the foreground classification~\cite{wood,eps}. The background semantic frequently appearing with the target foreground object carries suspicious information and results in ambiguous recognition, e.g., boat and lake (in (a)). The latter problem showcases (in (b)) that class activation maps prefer highlighting the discriminative semantic region instead of the entire object part~\cite{eps}. This issue manifests that the classifier uses less context-dependent information to explain its prediction due to the need for an in-depth understanding of class-level properties. These two issues motivate us to isolate the confusing background semantics from the classification process and learn more discriminative foreground features.

In this work, we analyze and conclude that the co-occurring problem of class activation maps resulted from disregarding the image background region in the classification training. However, existing background representation and feature separation methods~\cite{sipe,ICD} do not consider the foreground-background semantic discrepancy and thus fail to capture fine-granulate background semantics. To address these limitations, we propose a fine-grained background representation (FBR) method built upon contrastive learning. We recognize that the image background region contains a vast amount of task-irrelevant information and lacks descriptive semantics. Therefore, we model the background independently from the foreground. For every foreground class, we define a learnable prototype that encodes its essential features. In order to represent the image background, we specifically develop a primitive, negative-region-of-interest (NROI), capable of capturing the fruitful and diverse semantics of backgrounds. To address the co-occurring issue, we implement Fore-to-background contrast in a pixel-to-NROI manner to decouple the foreground objects effectively from the confusing background semantics and thus overcome the problem. Additionally, we tailor an active sampling method for weakly supervised semantic segmentation that selects negatives based on foreground semantic relationships and conduct intra-foreground contrastive learning. This way, we learn compacted foreground class features and obtain completer object regions.

In summary, our main contributions are three-fold:
\begin{itemize}
    \item We propose a simple FBR method to address the co-occurring and incomplete object region problem for weakly supervised semantic segmentation. 
\item We propose a fine-grained background primitive, dubbed NROI, to represent image background effectively and implement the fore-to-background contrastive learning to enhance class activation maps' ability to distinguish co-occurring background cues. Also, we introduce an active method to sample efficient foreground negatives and conduct intra-foreground contrastive learning to activate integral object regions.
\item Extensive experiments and evaluations in weakly supervised semantic and instance segmentation demonstrate our FBR approach can be used in different applications and is generalizable to various baseline architectures. In particular, our method achieves new state-of-the-art weakly supervised semantic segmentation performances on Pascal Voc 2012 and MS COCO 2014 test sets.
\end{itemize}

%% file: Content/Relatedwork_section.tex
\begin{table*}
    \centering
    \vspace{-10pt}
            \caption{We summarize related methods with informative keywords and highlight the differences between them and ours.}
    \begin{tabular}[t]{c<\centering |c<\centering |c<\centering }

    \Xhline{1pt}
       Method  & Publication Venue & Keywords \\
       \hline
        AdvCAM~\cite{advcam} & CVPR'21  & attribute maps, adversarial attack, image perturbations \\
        \hline
        ToCo~\cite{token} & CVPR'23  & token contrast, intermediate feature supervision, local-to-global consistency\\
        \hline
        
         C$^{2}$AM~\cite{ccam}&CVPR'22& class-agnostic activation map, contrastive learning, foreground-background  representations \\
         \hline
         PPC~\cite{wseg}& CVPR'22 & prototype learning, pixel-level supervision, cross-view semantic consistency\\
         \hline
        \rowcolor{Gray}Ours & -- & negative-region-of-interest, activate negative sampling, foreground-background contrastive learning\\
         \Xhline{1pt}

    \end{tabular}

    \label{tab:literature_review}
             \vspace{-15pt}
\end{table*}
\noindent\textbf{Image-level weakly supervised semantic segmentation.}  This task aims to generate reliable pseudo masks from class labels to guide the segmentation practice.  The main trend in this field is to produce complete object masks with class activation maps, yielding high-quality pseudo labels. 

Current studies~\cite{BECO,l2g} usually resort to getting more object information involved in the classification or using powerful backbone architectures, e.g., ViT~\cite{VIT}, to get more object regions activated. For example, RIB~\cite{RIB} presents a novel pooling method to reduce the information bottleneck during classification and forces classifiers to identify less discriminative regions of target class objects. ToCo~\cite{token} proposes contrasting the patch representations from different layers and the local-global features of the class token to scale up the activated region of target class objects.


Despite the advanced results achieved by these methods, some confusing background semantics are inevitably activated when expanding the object region, introducing noise pixels around semantic boundaries. In this work, we explicitly model the image foreground and background semantics and implement the fore-to-background and intra-foreground contrastive learning to suppress the effect of ambiguous background information, thus learning accurate object regions.

\noindent\textbf{Contrastive learning.}  The goal of this approach is to learn a similarity function in common feature space to pull views of similar data (positive) closer while pushing views of dissimilar ones (negative) apart. This way, we label each pixel based on the similarity measure and activate accurate object regions.

For instance, C$^{2}$AM~\cite{ccam} proposes the fore-background contrastive learning to generate the class-agnostic class activation maps. PPC~\cite{wseg} employs a prototype, namely the typical class features, to serve as the positive to perform the contrastive learning; besides, PPC adopts the “hardness” computation in~\cite{exploring_hard_sampling} to draw hard negatives. 

Nevertheless, most solutions~\cite{wseg,u2pl,reco} treat the image background equally as the foreground classes and optimize them jointly, ignoring the fact that the background content covers diverse object semantics. This nature makes the image background difficult to represent with a learnable prototype or features, resulting in inferior optimization effects. 

By contrast, we model the image background with the proposed novel primitive, negative-region-of-interst, to denote its complex semantics and perform the fore-to-background contrastive learning. This way,  we decouple the relationship between diverse background semantics and the foreground classes, addressing the co-occurring problem. Besides, we present an active sampling strategy to select the negative samples of foreground classes on the fly. Therefore, we accurately discriminate the intra-foreground relationships and learn integral object regions.\\  
\noindent\textbf{Technical comparison with related works:}: \refTab{tab:literature_review} compares our method with the aforementioned studies. We summarize each study with three keywords to introduce their main novelties and methodologies. To compare, our main contributions rely on the proposed negative-region-of-interest for diverse background semantic representation and the effective active sample method to facilitate the learning of foreground class prototypes. In Sec. \ref{sec:evaluation}, we conduct extensive experiments to verify the effectiveness of our method and demonstrate its benefits from different perspectives.

%% file: Content/Method_section.tex
In this part, we first clarify an underlying motivation of fine-grained background representation in weakly supervised semantic segmentation (WSSS). Next, we tailor the class activation maps (CAM)-based contrastive learning and propose optimizing the fore-to-background (FB) and the intra-foreground (IF) relationships.

\subsection{Motivation}
\label{sec:motivation}
In the image-level WSSS, we are given a training set $\mathcal{D}$, with data tuple $(\mathbf{x}, \mathcal{Y}) \in \mathcal{D}$, where each image $\mathbf{x}$ is associated with a class label $\mathcal{Y} = (y_1, y_2,...,y_C)$; $y_{c}=1$ denotes the presence of the foreground (FG) class $c$ ($1\leq c\leq C$) in $\mathbf{x}$ and 0 otherwise. While in the semantic segmentation task, we aim to learn a discriminative model (parameterized by neural networks) to approximate the conditional distribution $p(\mathbf{y}|\mathbf{x})$, where $\mathbf{y}\in \mathbb{R}^{(C+1)\times H \times W}$ ($H\times W$ denotes the spatial size) is the ground-truth semantic label that contains $C$ FG classes and a background (BG) class.

Existing WSSS solutions~\cite{cian, recam,psa,irn} utilize CAM (denoted with $\hat{\mathbf{y}}\in\mathbb{R}^{(C+1)\times H\times W}$) to approximate $p(\mathbf{y}|\mathbf{x})$ by learning a reliable semantic feature $f\in \mathbb{R}^{L\times H \times W}$ 
($L$ denotes the feature dimension) with the classification loss $L_{cls}$:
\begin{equation}
L_{cls} = -\frac{1}{C}\sum_{c=1}^C [y_{c}\log \sigma(\hat{s}_c)+(1-y_{c})\log (1-\sigma(\hat{s}_c))]
\label{eq:loss_cla},
\end{equation}
where $\sigma$ is softmax function and $\hat{s}_c$ is the classification score. We define CAM as a function that projects each pixel $i$'s feature $f_{i}$ with the parameter $\theta$ (the weights of the classifier) to the semantic label space $\hat{\mathbf{y}}_i\in \mathbb{R}^{C+1}$:
\begin{equation}
    CAM(f_{i},\theta)=\hat{\mathbf{y}}_i,
    \label{eq:cam_function}
\end{equation}
where $\hat{\mathbf{y}}_i$ is then normalized to a categorical distribution.

It is worth noting that $L_{cls}$ essentially learns $p(y_{c}|\mathbf{x}_{i})$,  i.e., the probability of the pixel $\mathbf{x}_{i}$ being assigned to the FG semantic space, excluding the BG. In the CAM generation step, we argue that the entire BG region is treated as a virtual class that is ignored, resulting in $f_{i}$ being vulnerable to the BG semantics, particularly co-occurring ones, and hence leading to classification ambiguity. This observation indicates the necessity of BG-oriented modeling. Also, with informative BG representations, we can further decouple the semantic correlation between the target objects~\cite{eps, sipe} and their nearby BG to better approximate the true $p(\mathbf{y}_{i}|\mathbf{x}_{i})$.

Our work is the first attempt to address the aforementioned limitations of CAMs by the fine-grained BG representation. Unlike existing approaches~\cite{reco,hard_contrastive,sipe} either using pixel features or prototypes to give an abstract BG representation, our key novelty resides in explicitly modeling the image BG with a novel fine-grained primitive and performing FB contrast to eliminate the BG confusion (Sec. \ref{sec:comparison}). Besides, we design an active negative sampling method to implement effective IF contrast, thus learning the compacted FG features to activate the complete object masks. We enhance CAMs and obtain more reliable seeds by optimizing these two relationships. 
\label{sec:overview}
\begin{figure*}[t]
\centering
\includegraphics[width=0.95\linewidth]{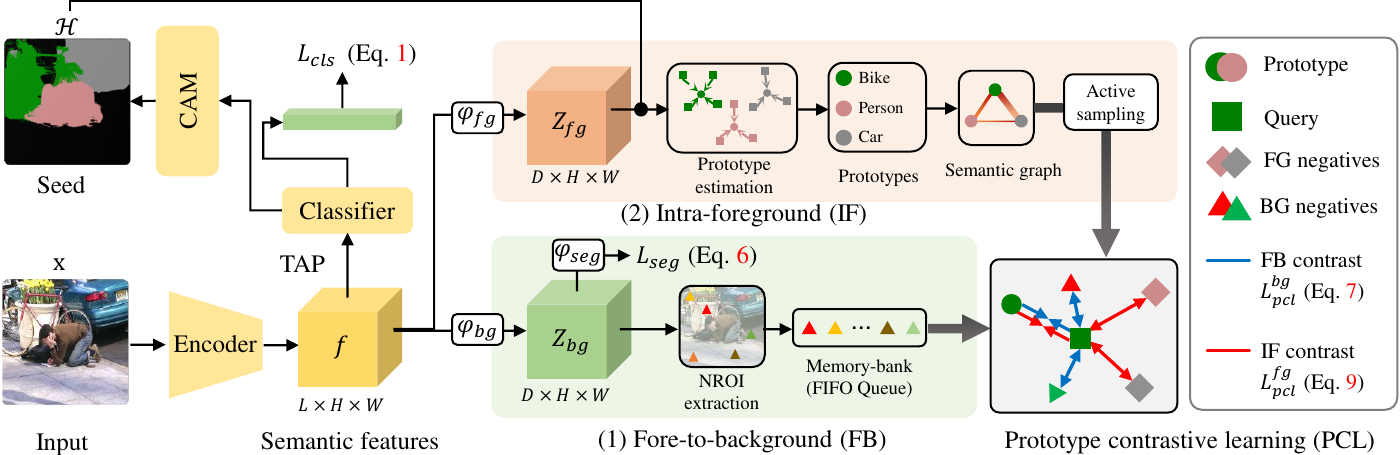}
   \caption{Architecture overview. A standard feature encoder trained with the classification loss $L_{cls}$ (with TAP~\cite{tap}) takes an input image $\mathbf{x}$ and generates the seed $\mathcal{H}$. We consider that image BG has a different semantic granularity from FG and add two projection heads, $\varphi_{fg}$ and $\varphi_{bg}$, model BG independently from FG to capture diverse BG information, and optimize two contrastive relationships: (1) fore-to-background and (2) intra-foreground. (1) enhances the semantic features $f$ in representing BG semantics with the proposed fine-grained primitive, namely NROIs. We compute FG prototypes and store NROIs in a memory bank. Besides, the auxiliary BG segmentation loss $L_{seg}$ is introduced. In (2), we present an active sampling strategy built upon the semantic graph to draw the FG negatives. The contrastive losses $L_{pcl}^{bg}$ for (1) and $L_{pcl}^{fg}$ for (2) pull the query closer to its prototype but push far from the FG and the BG negative keys, respectively.
   }
   \label{fig:architecture_overview}
   \vspace{-10pt}
\end{figure*}
\subsection{Our Method}
\label{sec:our_method}
\subsubsection{Contrastive learning setup for WSSS}
To begin with, we follow the pipeline in~\cite{pipeline} to generate the seed $\mathcal{H} \in \mathbb{R}^{ H \times W }$, except replacing the Global Average Pooling (GAP) layer with the Thresholded Average Pooling (TAP) layer~\cite{tap}, which averages only the above-threshold pixels in the semantic feature $f$ (introduced and ablated in the Supplementary). Additionally, we add a nonlinear projection head~\cite{reco,SimCLR}, $\varphi_{fg}$, encoding $f$ into a $D$-dimensional representation, $Z_{fg} \in \mathbb{R}^{ D \times H \times W }$, with the same spatial resolution as $\mathcal{H}$ (shown in \refFig{fig:architecture_overview}).\\
\textbf{FG prototype assignment:} According to the spatial location, we assign $\mathcal{H}$'s FG label information to the pixels in $Z_{fg}$. Following the setting in~\cite{region-aware,wseg}, for each FG class $c$ in the batch, we choose the top $N$ pixels with the high CAM scores and compute its prototype, $p_{c}$, as the weighted average of the pixel-level representation.
\begin{equation}
    p_{c}=\frac{\sum_{i\in\pi_{c}}\hat{\mathbf{y}}_{c,i}Z_{fg,i}}{\sum_{i\in\pi_{c}}\hat{\mathbf{y}}_{c,i}},
    \label{eq:prototype}
\end{equation}
where $\hat{\mathbf{y}}_{c}\in\mathbb{R}^{H\times W}$ denotes class $c$'s activation map, and $\pi_{c}$ is the spatial coordinate set of the top $N$ pixels in $\hat{\mathbf{y}}_{c}$.\\
\textbf{Query computation:} We build the query set, $Z^{q}_{c}$, for each appeared $c$ in the batch. Rather than querying all FG pixels, we consider CAM score as the certainty measure to determine $Z^{q}_{c}$ adaptively, enabling the contrastive loss focus on uncertain (below the threshold $\beta$, set to 0.4) pixels in $\mathcal{H}$:
\begin{equation}
    Z^{q}_{c}=\mathbbm{1}[\mathcal{H}=c]\cdot\mathbbm{1}[\hat{\mathbf{y}}_{c}<\beta]Z_{fg}.
    \label{eq:naive_query}
\end{equation}
\textbf{Prototype-based Contrastive Learning (PCL):} The standard contrastive loss~\cite{SimCLR,ProtoNCE} functions by encouraging the query $q\in Z_{c}^{q}$ to be similar to its positive keys and dissimilar to the negative key $z^n\in Z^n$. In this work, we employ the estimated FG prototypes $p_{c}\in P$ as the positives
and express the contrastive loss $L_{pcl}$ with:
\begin{equation}
\begin{split}
       L_{pcl}&=PCL(P,Z^{q}, Z^{n})\\
       &=\sum_{p_c\in P}\sum_{q\in Z_{c}^{q}}-\log \frac{e^{(q\cdot p_{c}/\tau)}}{e^{(q\cdot p_{c}/\tau)}+\sum_{z^{n}} e^{(q\cdot z^{n}/\tau)}},
           \end{split}
    \label{eq:basic_contrastive_loss}
\end{equation}
where $P=\{p_{c}\}_{c=1}^{C}$ and $Z^{q}=\{Z_{c}^q\}_{c=1}^{C}$ are the collections of prototypes and queries, $\tau$ and `$\cdot$' denote the temperature and dot product. We instantiate $Z^{n}$ from BG and FG regions and follow the function form in Eq. \ref{eq:basic_contrastive_loss} to respectively optimize the FB and IF contrastive relationships; loss functions are expressed in Eq. \ref{eq:bg_contrast} and Eq. \ref{eq:fg_contrast}.

\subsubsection{Fore-to-background (FB) Contrast}
\label{sec:fb_contrast}
In Sec. \ref{sec:motivation}, we conclude that the conditional BG distribution $p(\hat{\mathbf{y}}_{C+1}|\mathbf{x})$ can not be optimized by $L_{cls}$, and hence $f$ has weak BG description ability. Besides, image BG, unlike FG, does not have specific semantics and contains massive task-unrelated information; a single prototype, like Eq. \ref{eq:prototype}, is incapable of covering its high variance (discussed in Sec. \ref{sec:ablation}). Driven by these two concerns, we propose a fine-grained primitive, termed negative regions of interest (NROI), to comprehensively model the image BG that has a mixture of diverse semantics.\\
\textbf{NROI for BG representation:} Unlike existing approaches~\cite{SimCLR,hard_contrastive,ccam} that represent the FG and the BG semantics in a common space, we model the image BG independently to distinguish it from the FG well. Formally, we add another projection head, $\varphi_{bg}$ (shown in \refFig{fig:architecture_overview}), parallel with $\varphi_{fg}$, to look for the reliable BG representation from a different mapping: $\varphi_{bg}: f \rightarrow Z_{bg}, Z_{bg}\in \mathbbm{R}^{ D \times H \times W }$.

\begin{figure}[t]
\begin{center}
\includegraphics[width=8.3cm]{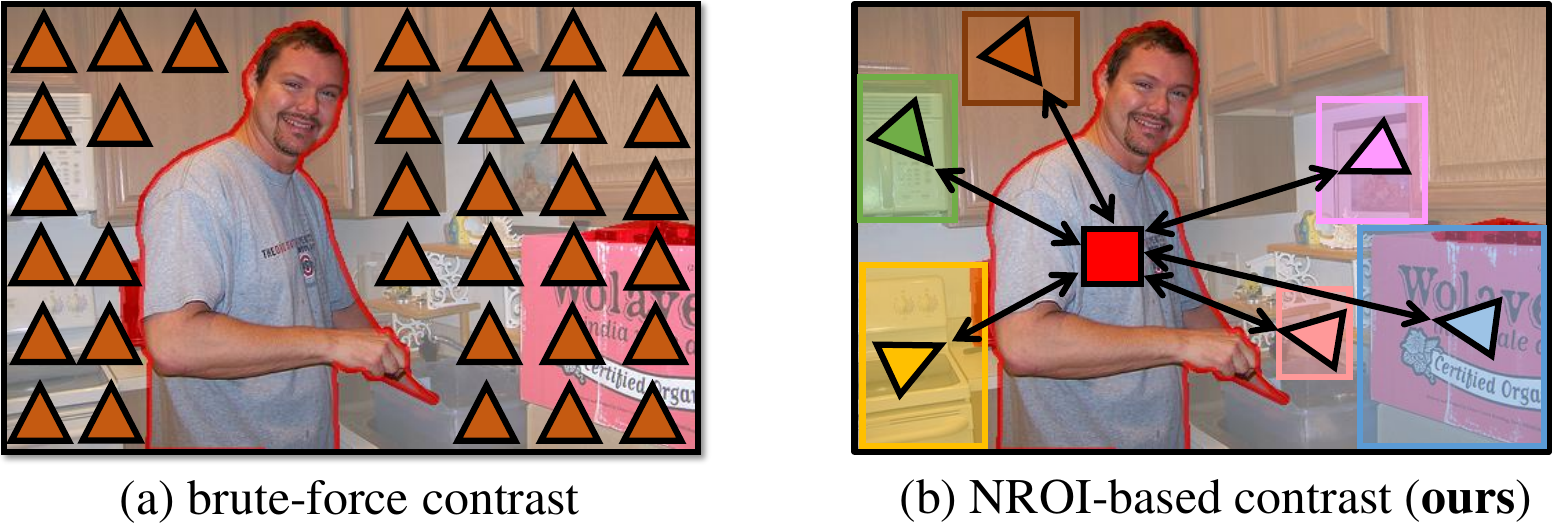}
\end{center}
 \vspace{-10pt}
   \caption{Conceptual illustration of negative-region-of-interest (NROI) for the FB contrast. The brute-force strategy (a) exhaustively compares FG queries (the red cropped part) with all BG pixels (triangles), which requires expensive computational resources and is susceptible to implausible labels. By contrast (b), we propose recognizing the fine-grained BG semantic, i.e., NROI. This example's NROIs (marked with different colors) contain the washing machine, closet, etc. In training, we implement FB contrastive learning by comparing queries (the red rectangle) against NROIs.}
   \label{fig:NROI_demonstration}
    \vspace{-10pt}
\end{figure}

A naive brute-force method (\refFig{fig:NROI_demonstration} (a)) is to use all BG features in $Z_{bg}$ to perform the optimization, i.e., pixel-to-pixel contrast, which would be time-consuming and computationally expensive. Also, the large-scale variation and high-complexity nature of the intra-BG region in training set $\mathcal{D}$ challenge us to develop an efficient representation method to denote its content. To this end, assuming that the image BG composes multiple semantics, we explore discovering fine-grained semantics and effectively denote them using NROI. Specifically, a group of $K$ NROIs $\{z_{bg}^{k}\}_{k=1}^{K}$ is used for BG description, where $k$ is the NROI indices with respect to the input $\mathbf{x}$.

We perform online clustering~\cite{rethink_prototype} for NROI determination. We get the BG features (with the spatial information in $\mathcal{H}$) from the masked $Z_{bg}$ and map them to $K$ clusters with K-means. Intuitively, clustering~\cite{ProtoNCE,rethink_prototype} imposes an inductive bias~\cite{rethink_prototype} that image BG consists of multiple semantics, and it thereby enables the model to discover the discriminative pixel groups, i.e., semantics. Hence, the NROIs $\{z_{bg}^{k}\}_{k=1}^{K}$ of each image, defined as the cluster centroids, are the typical representations of the BG semantics (in \refFig{fig:NROI_demonstration} (b)).

\noindent\textbf{BG memory bank:} We construct a queue-based memory bank~\cite{memo_bank}, $Z_{bg}^{n}$, to store NROIs, and set it as fixed storage for spatial and computational efficiency. As shown in \refFig{fig:architecture_overview}, the bank is updated at each training step with the extracted NROIs, and then we use the BG negative keys randomly sampled from $Z_{bg}^{n}$ to contrast against FG queries.

\noindent\textbf{Auxiliary BG segmentation.} 
Unlike prior studies~\cite{SimCLR,u2pl}, FBR adopts two projection heads to perform the contrastive learning, which brings a risk of a homogeneous representation between $Z_{fg}$ and $Z_{bg}$. To avoid this trivial case, we formulate a learning objective to the BG representation $Z_{bg}$, distinguishing it from $Z_{fg}$ and enhancing its BG discrimination ability. Specifically, we introduce binary segmentation as an auxiliary task, empirically considering pixels in $\mathcal{H}$ with a low-summed FG activation value~\cite{psa} (smaller than 0.05) as the pseudo BG labels (termed $M$), and feed $Z_{bg}$ (after batch normalization) into the BG predictor, $\varphi_{seg}$:
  \begin{equation}
    L_{seg}=BCE(\varphi_{seg}(Z_{bg}),M),
    \label{eq:bg_seg_loss}
\end{equation}
where BCE is a binary cross entropy loss.\\
\textbf{Pixel-to-NROI contrast:} With the query sets (Eq. \ref{eq:naive_query}) and the BG memory bank, we give the FB contrastive loss as:
\begin{equation}
    L_{pcl}^{bg}=PCL(P,Z^{q}, Z_{bg}^{n}).
    \label{eq:bg_contrast}
\end{equation}
This pixel-to-NROI contrast maximizes the agreement between the queries and their belonging prototype while minimizing the agreement with the BG semantic, i.e., NROIs.
\subsubsection{Intra-foreground (IF) contrastive learning}
\label{sec:intra_foreground}
This part presents an active negative sampling method to select FG negative keys and conduct effective IF contrast. 

\noindent\textbf{Active negative sampling:} We first define the query class $c$'s full negative set, $z_{c}^{n}$, which contains all FG pixels that do not belong to $c$: i,e., $z_{c}^{n}=\mathbbm{1}[(\mathcal{H}\neq c)\cap(\mathcal{H}\neq C+1)]Z_{fg}$. However, contrasting the query against all samples in $z_{c}^{n}$ is computationally costly. Moreover, contrastive learning may be ineffective or even degenerate an overall performance due to the implausible label information of the seed $\mathcal{H}$.

Inspired by the recent semi-supervised semantic segmentation study~\cite{reco}, we propose actively drawing negatives from $z_{c}^{n}$ and optimizing only with the selected samples to overcome the above limitations. For each batch, we compute a graph $G\in \mathbb{R}^{C\times C}$, where nodes and edges stand for the occurring classes and their relative semantic relations:
\begin{equation}
    G[i,j]=Sim(p_{i},p_{j}), \text{where } i,j \le C, \text{and }  i\neq j.
    \label{eq:graph_computation}
\end{equation}
Here, $Sim$ is the cosine similarity. Unlike~\cite{reco}, we exclude the BG class from the graph and use the semantic distance between FG prototypes, rather than between the mean features, to measure the pair-wise relationship $G[i,j]$.

For each query class $c$, we turn its relationships in $G$ against negative classes into a distribution by softmax: $\frac{\exp(G[c, i])}{\sum_{j\le C,j\ne c} \exp(G[c,j])}$. We sample keys of each negative class $i$ from $z_{c}^{n}$ based on the distribution. Intuitively, this step performs a non-uniform sampling on $z_{c}^{n}$, drawing more samples from the classes that are semantically similar to $c$, while drawing fewer from dissimilar ones. It enables the classifier to learn compacted FG features and an accurate decision boundary by improving the discrimination ability regarding the confusing, negative classes. 

\noindent\textbf{Pixel-to-prototype contrast:} Similar to Eq. \ref{eq:bg_contrast}, we formulate the IF contrastive loss as:
\begin{equation}
    L_{pcl}^{fg}=PCL(P,Z^{q}, Z_{fg}^{n}),
    \label{eq:fg_contrast}
\end{equation}
where $Z_{fg}^{n}=\{z_{c}^{n}\}_{c=1}^{C}$. This pixel-to-prototype contrast learns compacted FG features by pulling the queries close to their prototype and pushing different classes far away.
\subsection{Overall objective}
As illustrated in \refFig{fig:architecture_overview}, our FBR method can be integrated into existing WSSS solutions to obtain better seeds. We add two projection heads $\varphi_{fg}$ and $\varphi_{bg}$ after the encoder, mapping the semantic feature $f$ into two high-dimensional representations $Z_{fg}$ and $Z_{bg}$, and then implement the FB and the IF contrastive learning. The overall loss is expressed:
\begin{equation}
    L=L_{cls}+ L_{pcl}+L_{seg},
    \label{eq:full_loss_formulation}
    \end{equation}
where $L_{pcl}=\lambda_{1}\cdot L_{pcl}^{bg}+\lambda_{2}\cdot L_{pcl}^{fg}$, jointly optimizing these two constrative relationships using corresponding loss weights $\lambda_{1}$ and $\lambda_{2}$. Note that $\varphi_{fg}$ and $\varphi_{bg}$ are only applied during training and discarded in the inference phase.


%% file: Content/Experiment_section.tex
\label{sec:evaluation}
We first ablate FBR to test its effectiveness and apply it to existing models to get state-of-the-art (SOTA) results.  
\subsection{Setup}
\noindent\textbf{Datasets \& evaluation metrics.} We experiment on two benchmarks, Pascal Voc 2012~\cite{pascal} and MS COCO 2014 \cite{coco}. The former contains 20 object classes, with 10,582 images for training, 1,449 images for validation, and 1,456 for testing. MS COCO 2014 has 80 labeled classes, 80,781 training and 40,321 validation images. We evaluate the generated pseudo labels and the segmentation results with their ground-truth segmentation labels. Both experiments are evaluated with mean intersection over union (mIoU). Besides, we further explore the effectiveness of our method on weakly supervised instance segmentation~\cite{boxinst,boxlevelset} (WSIS, with box-level annotations). We conduct experiments on MS COCO 2017 dataset, which has 115K images for training and 5K evaluation images. During inference, we report AP, AP$_{50}$, AP$_{75}$ (averaged precision over different IoU thresholds) for the instance segmentation performance evaluation.\\
\noindent\textbf{Implementation details.} We set PPC~\cite{wseg} as the baseline and perform ablation experiments on the proposed FB and IF contrastive learning to test their effectiveness and compare them with peer studies. Next, we add the full method to various WSSS models~\cite{AMN,wseg,seam} and experiment on the aforementioned datasets to show our FBR's generality and superiority. In training, we set the cluster number $K$ to 8. We implement the projection head $\varphi_{fg}$ and $\varphi_{bg}$ with a $1\times1$ convolutional layer followed by ReLU and set $D$ to 128. We return 256 negative keys for every query and set $\tau$ in Eq. \ref{eq:bg_contrast} and Eq. \ref{eq:fg_contrast} to 0.5 and 0.1. For semantic segmentation practice, we adopt DeepLab-v2-ResNet101~\cite{DeepLabv2} as the backbone. In box-level WSIS, we deploy FBR on BoxInst~\cite{boxinst}, set the feature dimensionality $D$ to 64, and keep other hyperparameters unchanged. To reduce the computational overhead, all images are resized to have their shorter side in the range [480, 640] during training. More details are reported in the Supplementary.

\subsection{Overall results}

\label{sec:comparison}
In this section, we compare the proposed FBR method with existing WSSS studies regarding the accuracy of generated pseudo labels and yielding semantic segmentation results. Besides, we extend FBR to WSIS tasks to show its benefits.

\begin{figure*}[t]
\begin{center}
\includegraphics[width=17.4cm]{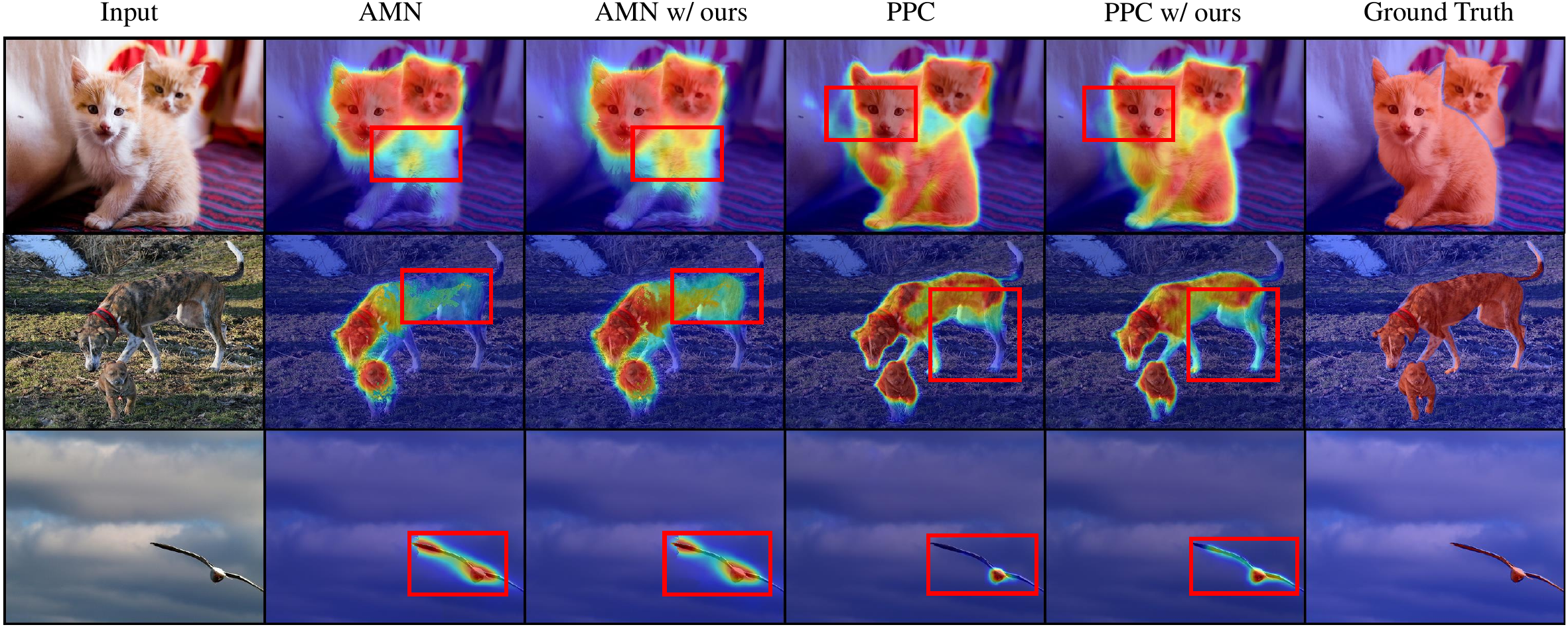}
\end{center}
\vspace{-5pt}
   \caption{Example results of CAMs on Pascal Voc 2012 train set. From left to right: input images, results of AMN, results of AMN w/ ours, results of PPC, results of PPC w/ ours and the ground truth. The red boxes highlight the refined details.
   }
   \label{fig:cam_comparison_visualization}
    \vspace{-10pt}
\end{figure*}
\begin{figure*}[t]
\begin{center}
\includegraphics[width=17.4cm]{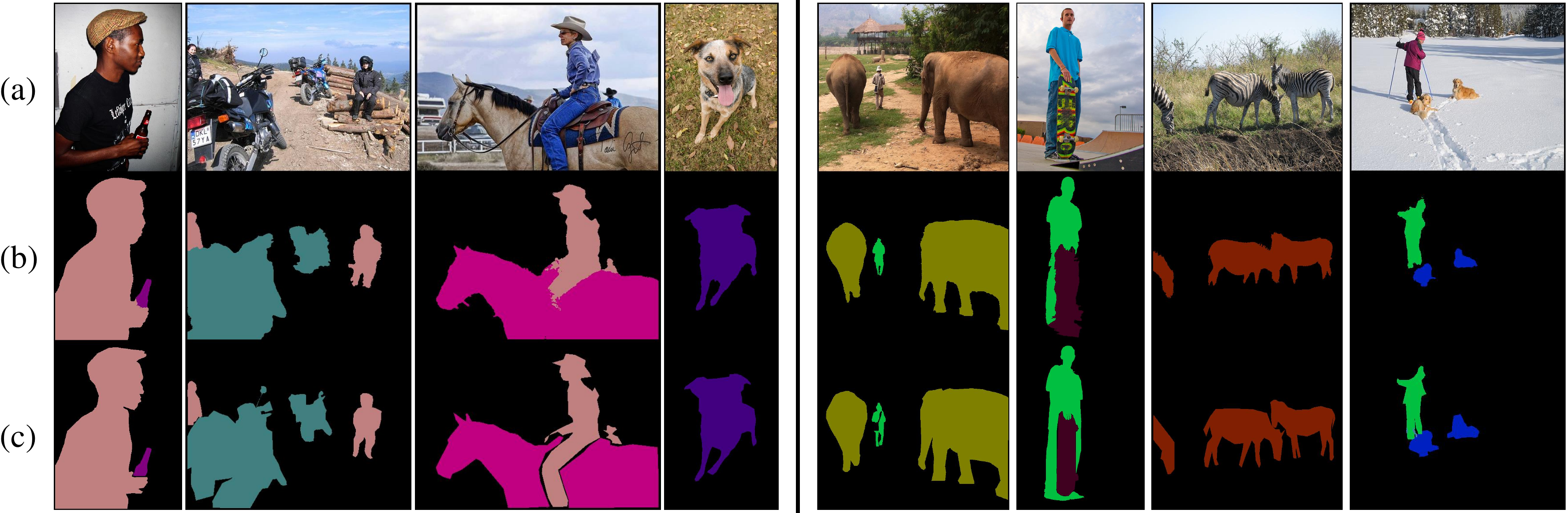}
\end{center}
\vspace{-5pt}
   \caption{Qualitative semantic segmentation results. The left figures are results from Pascal Voc 2012 val set, and the right ones are from MS COCO 2014 val set. (a) Input images, (b) Ours, (c) Ground truth.
   }
   \label{fig:final_segmentation_result}
    \vspace{-10pt}
\end{figure*}

\begin{figure*}[t]
\begin{center}
\includegraphics[width=17.4cm]{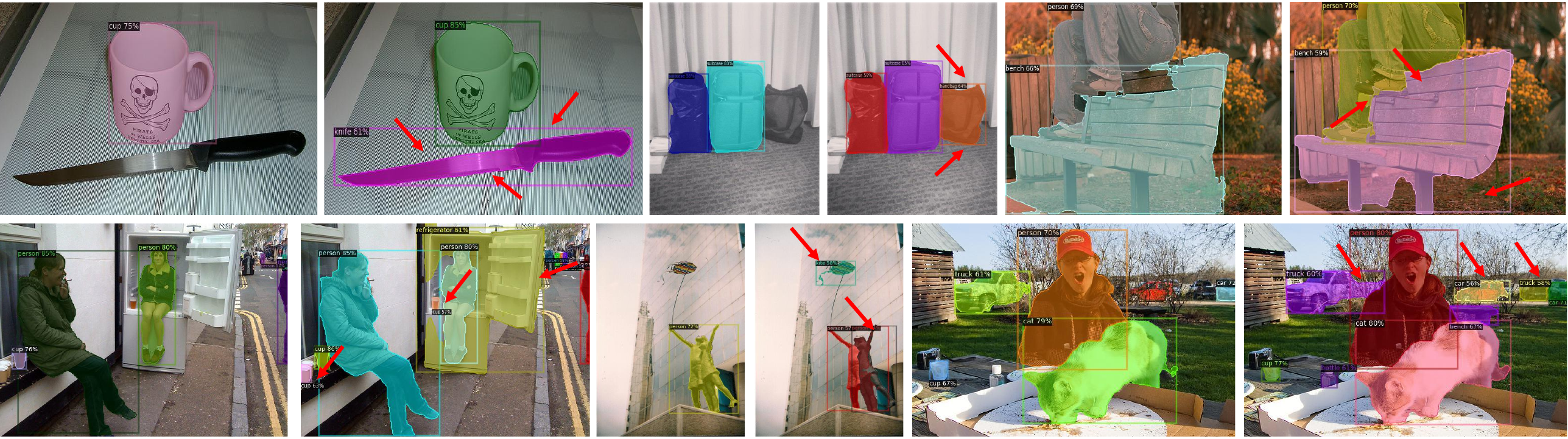}
\end{center}
 \vspace{-10pt}
   \caption{Qualitative instance segmentation results on MS COCO 2017 val set. Note that example pairs are from BoxInst~\cite{boxinst} (left images) and ours (BoxInst w/ FBR, right images); we see that FBR substantially fines the mask predictions. 
   }
   \label{fig:final_instance_visualization}
   \vspace{-10pt}
\end{figure*}

\noindent\textbf{Results of pseudo labels:} In this part, we evaluate the quality of the pseudo-segmentation masks. We add FBR on three representative baselines, PPC~\cite{wseg} that use saliency map (I+S) in training, SEAM~\cite{seam} and AMN~\cite{AMN} (I) that do not use, to manifest our approach's generality. Note that SEAM and PPC require cross-view inputs, while AMN does not. \refTab{tb:overall_pseudo_labels} reports the comparison performance. In Seed results, FBR achieves 7.2\%/1.7\%/0.7\% mIoU improvements on SEAM/PPC/AMN. These gains are almost maintained after employing CRF (Conditional Random Field), IRN~\cite{irn}, or PSA~\cite{psa} to refine the seeds and obtain the pseudo segmentation labels (\textbf{Mask}). Overall, our method achieves SOTA performance on Pascal Voc 2012 train set, surpassing the best-known WSSS study by 1.8\% mIoU (75.9\% $vs.$ 74.1\%).

\refFig{fig:cam_comparison_visualization} compares the qualitative results of CAMs. Note that FBR greatly helps the baselines activate more object regions (the dog and cat example on AMN~\cite{AMN} column) and enhances their background discrimination ability and thus learns more accurate object boundaries (see the dog example on PPC).

\begin{table}
\small
    \caption{mIoU (\%) comparison of the seeds, seeds w/ CRF, and pseudo masks (\textbf{Mask}) on Pascal Voc 2012 train set. Throughout the paper, bolded and underlined represent the SOTA and the second-best SOTA results, respectively.}
    \label{tb:overall_pseudo_labels}
    \centering
    \begin{tabular}[t]{m{0.2\textwidth}<\centering m{0.06\textwidth}<\centering m{0.06\textwidth}<\centering m{0.06\textwidth}<\centering}

\Xhline{1pt}   
\textbf{Method} &\textbf{Seed}& \textbf{+CRF} & \textbf{Mask} \\
\Xhline{1pt} 
Refine with PSA \cite{psa}: &&&\\
SEAM \scriptsize{CVPR '20}~\cite{seam}&55.4&56.8&63.6\\
\rowcolor{Gray}
Ours (SEAM-based)&62.6\tiny{+7.2}&65.3\tiny{+8.5}&69.9\tiny{+6.3}\\
RIB \scriptsize{NeurIPS '21}~\cite{RIB}&56.5&62.9&68.6\\
EPS \scriptsize{CVPR '21}~\cite{eps}&69.4&71.4&71.6\\
RCA \scriptsize{CVPR '22}\cite{rca}&-&-&\underline{74.1}\\
PPC \scriptsize{CVPR '22}\cite{wseg} &\underline{70.5}&\underline{73.3}&73.3\\
\hline
\rowcolor{Gray}
Ours (PPC-based)&\textbf{72.2\tiny{+1.7}}&\textbf{75.5\tiny{+2.2}}&\textbf{75.9\tiny{+2.6}}\\
\hline
Refine with IRN \cite{irn}: &&&\\
MCT \scriptsize{CVPR '22}\cite{mctformer}&61.7&-&69.1\\
CLIMS \scriptsize{CVPR '22} \cite{clims}&56.6&-&70.5\\
W-OoD \scriptsize{CVPR '22}\cite{wood} &59.1&65.5&72.1\\
AMN \scriptsize{CVPR '22}\cite{AMN}&\underline{62.2}&-
&72.2\\
ACR \scriptsize{CVPR '23}\cite{ACR}&60.9&65.9
&72.3\\
BECO \scriptsize{CVPR '23}\cite{BECO}&\-&-
&\underline{73.0}\\
\hline
\rowcolor{Gray}
Ours (AMN-based)&\textbf{62.9\tiny{+0.7}}&-&\textbf{73.1\tiny{+0.1}}\\\Xhline{1pt} 
    \end{tabular}
 \vspace{-10pt}
\end{table}

\begin{table}
\small
    \caption{Segmentation results (mIoU \%) comparisons with other SOTA studies on Pascal Voc 2012 validation (\textbf{Val.}) and test (\textbf{Test}) set. The supervisions (\textbf{Sup.}) used in the training include the image-level labels (\textbf{I}) and salience maps (\textbf{S}).}
    \label{tab:overall_pascal_segmentation}
    \centering
    \begin{tabular}[t]{c<\centering c<\centering c<\centering c<\centering c<\centering}
    \Xhline{1pt}   
\textbf{Method} &\textbf{Sup.}& \textbf{Backbone}&\textbf{Val.} & \textbf{Test} \\
    \Xhline{1pt}  
SEAM \scriptsize{CVPR '20}~\cite{seam}&I&ResNet38&64.5&65.7\\
ReCAM \scriptsize{CVPR '22}~\cite{recam}&I&ResNet101&68.4&68.2\\
SIPE \scriptsize{CVPR '22}~\cite{sipe}&I&ResNet101&68.8&69.7\\
W-OoD \scriptsize{CVPR '22}~\cite{wood}&I&ResNet38&70.7&70.1\\
AMN \scriptsize{CVPR '22}~\cite{AMN}&I&ResNet101&70.7&70.6\\
VIT-PCM \scriptsize{ECCV '22}~\cite{VIT-PCM}&I&ResNet101&70.3&70.9\\
SBCE \scriptsize{ECCV '22}~\cite{SBCE}&I&ResNet101&70.0&71.3\\
AEFT \scriptsize{ECCV '22}~\cite{AEFT}&I&ResNet38&70.9&71.7\\
BECO \scriptsize{CVPR '23}~\cite{BECO}&I&ResNet38&\underline{72.1}&71.8\\
ToCo \scriptsize{CVPR '23}~\cite{token}&I&VIT-B~\cite{VIT}&71.1&72.2\\
ACR \scriptsize{CVPR '23}~\cite{ACR}&I&ResNet38&\textbf{72.4}&\underline{72.4}\\
\hline
\rowcolor{Gray}
Ours (SEAM-based)&I&ResNet38&68.9&69.3\\
\rowcolor{Gray}
Ours (AMN-based)&I&ResNet101&71.8&\textbf{73.2}\\
\hline
MCT \scriptsize{CVPR '22}~\cite{mctformer}&I+S&ResNet38&71.9&71.6\\
L2G \scriptsize{CVPR '22}~\cite{l2g}&I+S&ResNet101&72.1&71.7\\
ReCAM \scriptsize{CVPR '22}~\cite{recam}&I+S&ResNet101&71.8&72.2\\
RCA \scriptsize{CVPR '22}~\cite{rca}&I+S&ResNet38&72.2&72.8\\
SBCE \scriptsize{ECCV '22}~\cite{SBCE}&I+S&ResNet101&71.8&73.4\\
PPC \scriptsize{CVPR '22}~\cite{wseg}&I+S&ResNet101&\underline{72.6}&\underline{73.6}\\
\Xhline{1pt}
\rowcolor{Gray}
Ours (PPC-based)&I+S&ResNet101&\textbf{74.2}&\textbf{74.9}\\
   \Xhline{1pt}

    \end{tabular}
 \vspace{-10pt}
\end{table}

\begin{table}
\small
    \caption{Semantic segmentation (mIoU \%) comparisons with other WSSS studies on COCO 2014 validation set.}
    \label{tab:overall_coco_segmentation}
    \centering
    \begin{tabular}[t]{c<\centering c<\centering c<\centering c<\centering}
\Xhline{1pt}  
\textbf{Method} &\textbf{Sup.}& \textbf{Backbone} & \textbf{Val.} \\
\hline

SIPE \scriptsize{CVPR '22}\cite{sipe}&I&ResNet38&43.6\\
RIB \scriptsize{NeurIPS '21}\cite{RIB}&I+S&ResNet101&43.8\\
L2G \scriptsize{CVPR '22}\cite{l2g}&I+S&ResNet101&44.2\\
AMN \scriptsize{CVPR '22}\cite{AMN}&I&ResNet101&44.7\\
AEFT \scriptsize{ECCV '22}~\cite{AEFT}&I&ResNet38&44.8\\
ReCAM \scriptsize{CVPR '22}\cite{recam}&I&ResNet101&45.0\\
VIT-PCM \scriptsize{ECCV '22}~\cite{VIT-PCM}&I&ViT-B/16~\cite{VIT}&45.0\\
BECO \scriptsize{CVPR '23}~\cite{BECO}&I&ResNet101&45.1\\
ACR \scriptsize{CVPR '23}~\cite{ACR}&I&ResNet38&\underline{45.3}\\
\hline
\rowcolor{Gray}
Ours (AMN-based)&I&ResNet101&\textbf{45.6}\\
\Xhline{1pt} 

    \end{tabular}

  \vspace{-10pt}
\end{table}

\begin{table}
\small
    \caption{Instance segmentation comparisons (AP \%) with other WSIS studies (box-level) on COCO 2017 validation set.  }
    \label{tab: coco_wsis_segmentation}
    \centering
    \begin{tabular}[t]{c<\centering c<\centering c<\centering c<\centering c<\centering c<\centering c<\centering c<\centering c<\centering}
\Xhline{1pt}  
\textbf{Method} &\textbf{Backbone} & \textbf{AP}& \textbf{AP$_{50}$}& \textbf{AP$_{75}$}\\
\hline
DiscoBox~\cite{discobox} \scriptsize{ICCV '21}&ResNet50&30.2&52.1&30.7\\
Box2Mask-C~\cite{boxlevelset} \scriptsize{ECCV '22}&ResNet101&33.5&56.9&34.2\\
BoxInst~\cite{boxinst} \scriptsize{CVPR '21}&ResNet50&30.9&53.3&31.2\\
BoxInst~\cite{boxinst} \scriptsize{CVPR '21}&ResNet101&32.1&55.3&32.4\\
\rowcolor{Gray}Ours (BoxInst-based)& ResNet50&\textbf{32.9}&56.6&33.4\\
\rowcolor{Gray}Ours (BoxInst-based)& ResNet101&\textbf{34.1}&57.7&34.9\\
\hline
\Xhline{1pt} 
 \vspace{-10pt}
    \end{tabular}

\end{table}

\noindent\textbf{Results of semantic segmentation.} We conduct semantic segmentation practices with DeepLabV2-ResNet101~\cite{DeepLabv2} and follow the training settings of existing implementations \footnote{https://github.com/YudeWang/deeplabv3plus-pytorch.git} \footnote{https://github.com/kazuto1011/deeplab-pytorch.git}. \refTab{tab:overall_pascal_segmentation} reports the mIoU results on Pascal Voc 2012 validation set and the test set, and compares our FBR method with recent WSSS studies. Training under the same setting, we improve SEAM and AMN by 3.6\% and 1.7\% mIoU on the test set of Pascal Voc, yielding the SOTA performance in the image-level (I) setting (72.4\%$\rightarrow$ 73.2\% mIoU \footnote{http://host.robots.ox.ac.uk:8080/anonymous/30LARO.html}). 

Besides, PPC~\cite{wseg} equipped with our FBR method achieves 74.2\% mIoU and 74.9\% mIoU \footnote{http://host.robots.ox.ac.uk:8080/anonymous/BHSCOK.html} on Pascal Voc 2012 Val and test sets, exceeding all (I+S) WSSS studies. Moreover, FBR achieves 45.6\% mIoU on MS COCO 2014 val set (\refTab{tab:overall_coco_segmentation}), obtaining the SOTA result. \refFig{fig:final_segmentation_result} presents example segmentation results on both benchmarks. 

\noindent\textbf{Results of instance segmentation.} In \refTab{tab: coco_wsis_segmentation}, we test FBR on BoxInst~\cite{boxinst} and report weakly supervised instance segmentation results on MS COCO 2017. Training under the same settings, FBR improves the baseline model by 2.0\% AP no matter the backbone models, surpassing Box2Mask~\cite{boxlevelset} by 0.6\% (33.5\%$\rightarrow$34.1\%). \refFig{fig:final_instance_visualization} shows the example results.

\begin{table}
\small
    \caption{Ablation study (on PPC~\cite{wseg}) in terms of seed generation on Pascal Voc 2012 train set. $\dagger$ denotes excluding the background in the training. $\ddagger$ denotes adopting the auxiliary background segmentation on $\dagger$. \textbf{FB:} fore-to-background contrast. \textbf{IF:} intra-foreground contrast.}
    \label{tab:abl_fg_bg}
    \centering
    \begin{tabu}[t]{m{1.5cm}<\centering m{1.5cm}<\centering m{1.5cm}<\centering m{1.5cm}}
    \tabucline[1pt]{-}
    \textbf{Baseline} &\textbf{FB}&\textbf{ IF}&\textbf{mIoU(\%)} \\
    \tabucline[1pt]{-}
    \checkmark&&&73.3\\
\text{$\checkmark^{\dagger}$}&&&73.6\tiny{+(0.3)}\\
\text{$\checkmark^{\dagger}$}&\checkmark&&74.7\tiny{+(1.4)}\\
\text{$\checkmark^{\ddagger}$}&\checkmark&&75.0\tiny{+(1.7)}\\
\checkmark&&\checkmark&73.9\tiny{+(0.6)}\\
\text{$\checkmark^{\ddagger}$}&\checkmark&\checkmark&\textbf{75.5\tiny{+(2.2)}}\\
    \tabucline[1pt]{-}
  
    \end{tabu}
  \vspace{-10pt}
    
\end{table}

\begin{figure}[t]
\includegraphics[width=8.3cm]{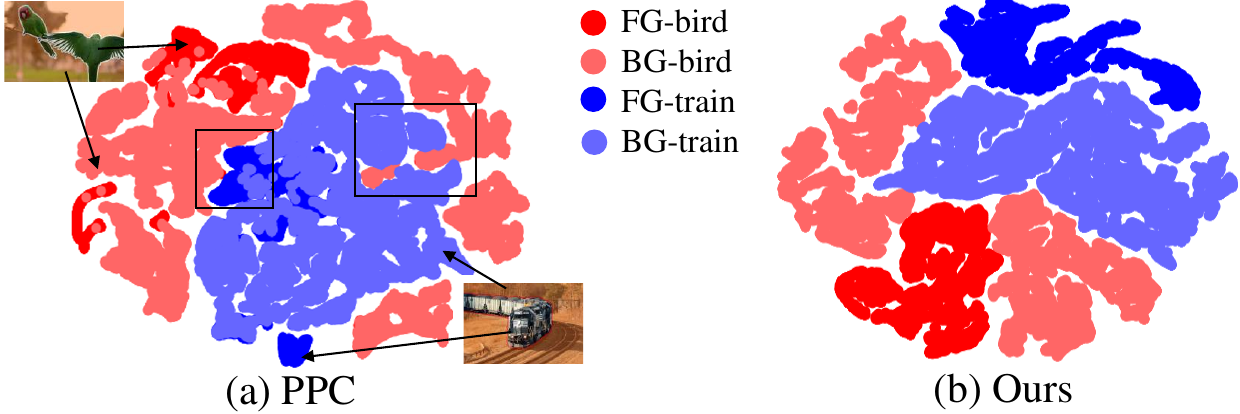}
   \caption{Background-foreground feature visualization via t-SNE~\cite{tsne}. (a) foreground features are confused with background information (the cropped regions). (b) FB contrast learns compacted background features by fine-grained recognition and well separates background and foreground features. }
                \label{fig:bg_tsne}
               \vspace{-15pt}
\end{figure}

\begin{figure}[t]
\includegraphics[width=8.3cm]{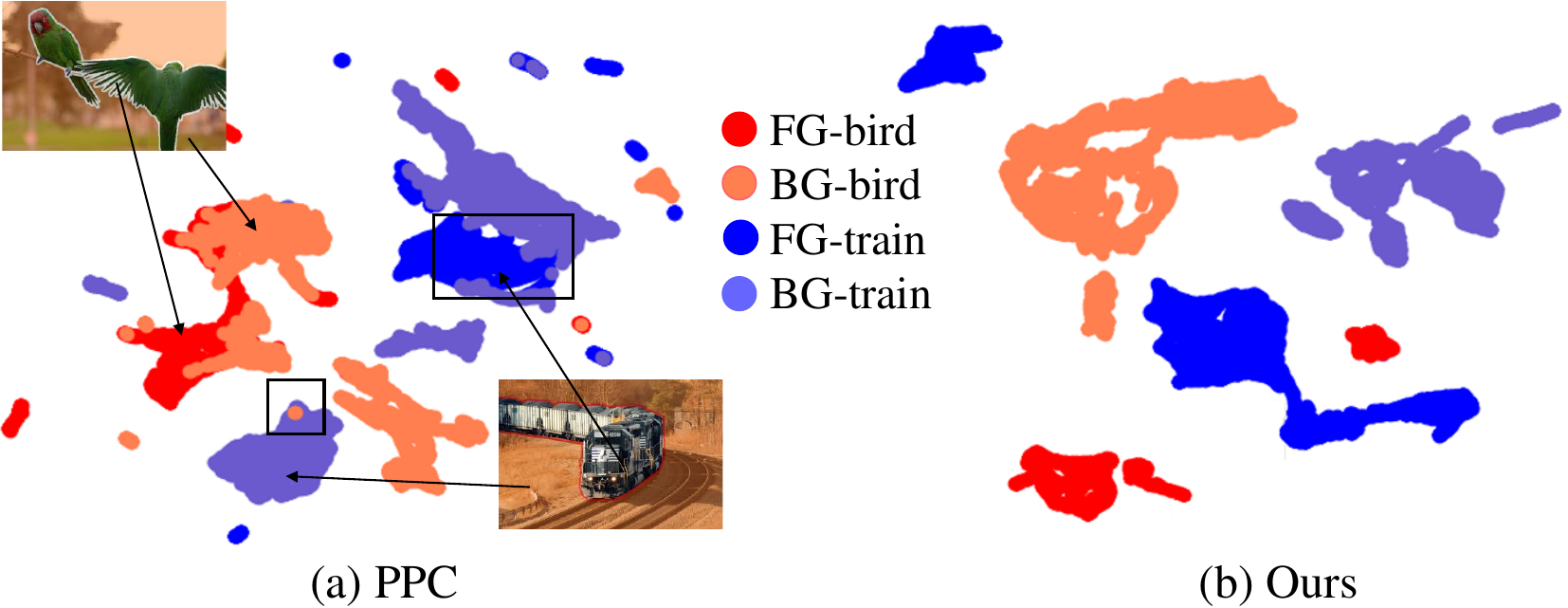}
   \caption{UMAP~\cite{umap} visualization for background-foreground feature comparison. }
                \label{fig:bg_umap}
                \vspace{-15pt}
\end{figure}

\begin{figure}[t]
\begin{center}
\includegraphics[width=8.3cm]{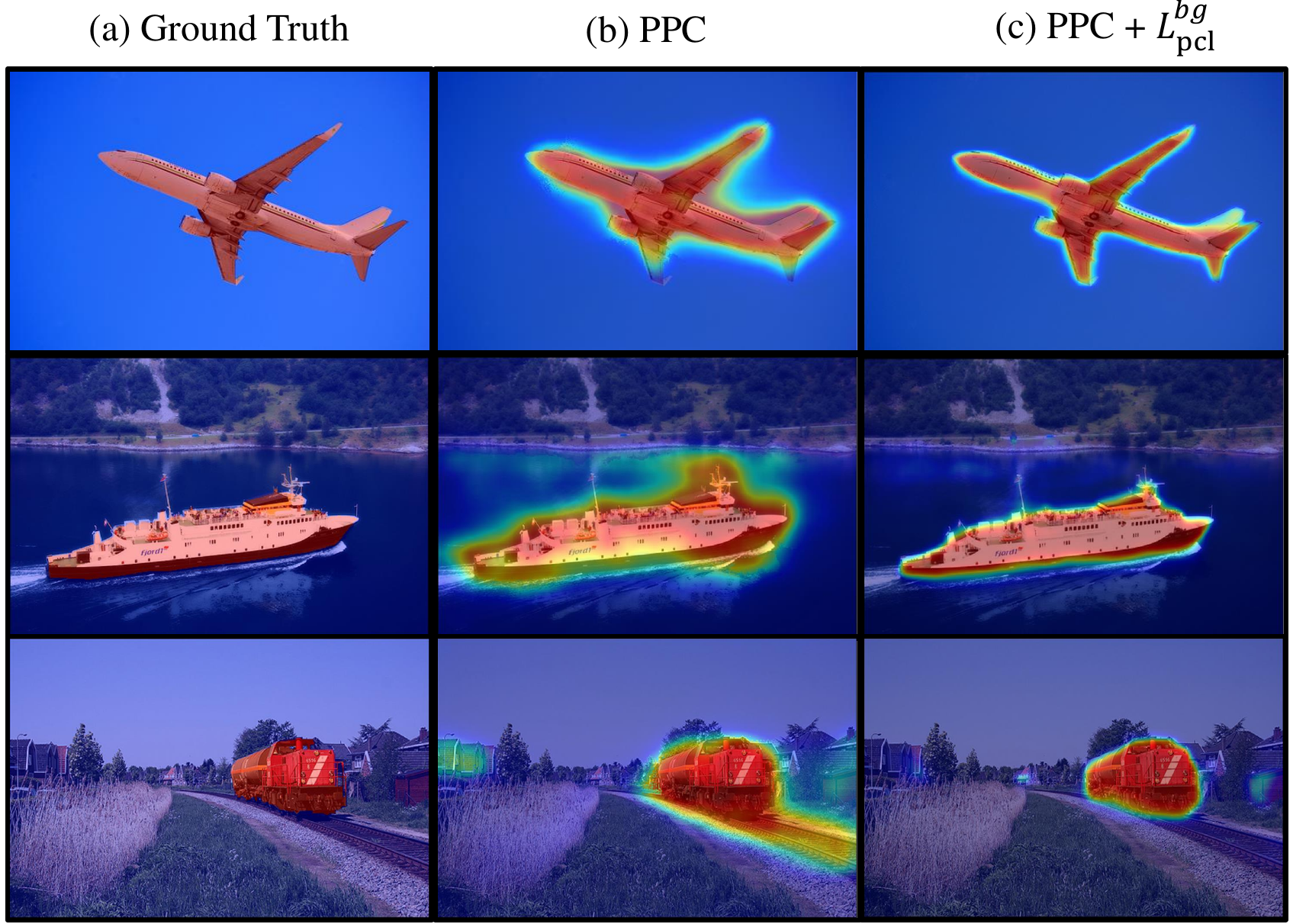}
\end{center}
 \vspace{-10pt}
   \caption{Effect of FB contrast (i.e., $L_{pcl}^{bg}$) on PPC~\cite{wseg}. 
   }
   \label{fig:co_occurring_refinement}
    \vspace{-10pt}
\end{figure}
\begin{figure}[t]
\begin{center}
\includegraphics[width=8.3cm]{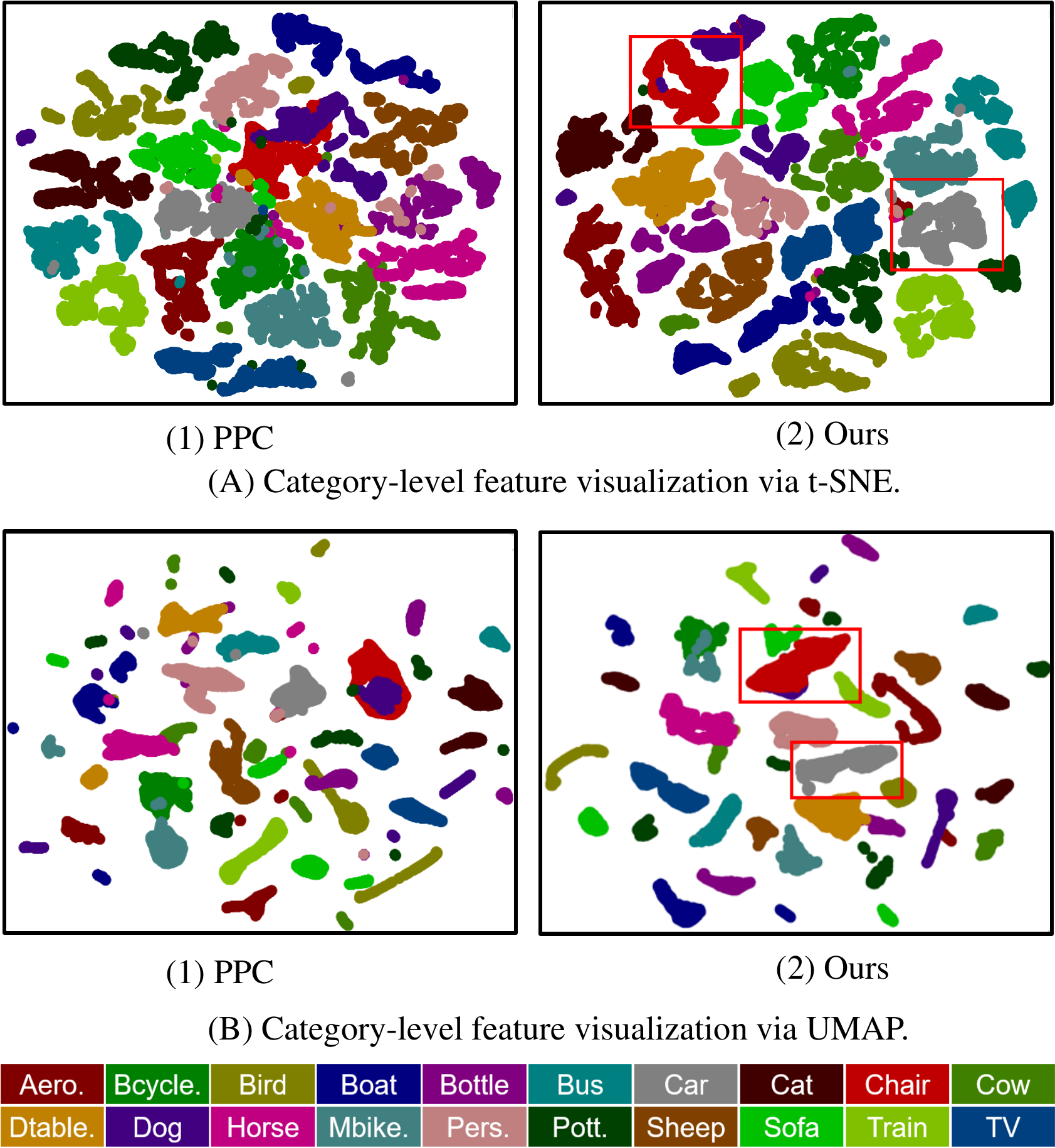}
\end{center}
 \vspace{-10pt}
   \caption{Visualization of semantic features via (A) t-SNE and (B) UMAP. \textbf{Left}: PPC~\cite{eps}. \textbf{Right}: PPC w/ $L_{pcl}^{fg}$.
   }
   \label{fig:fg_tsne}
    \vspace{-10pt}
\end{figure}
\begin{table}[h]
\small
        \caption{NROI evaluation. \textbf{Left:} w/o and w/ $\varphi_{bg}$ mean that background features to obtain NROIs are from $Z_{fg}$ and $Z_{bg}$, respectively. \textbf{Right:} Strategy comparison. Ours: pixel-to-NROI. Brute-force: pixel-to-pixel.}
  \label{tab:ablation_NROI}
\centering
  \begin{minipage}[t]{4cm}
  \small
    \centering
    \begin{tabular}{ c c }
      \toprule
      \textbf{Method} & \textbf{mIoU(\%)} \\
      \toprule
      w/o $\varphi_{bg}$ & 74.0\\
      w/ $\varphi_{bg}$  & \textbf{74.7}\\
      \toprule
    \end{tabular}

  \end{minipage}
  \begin{minipage}[t]{4cm}
  \small
    \centering
    \begin{tabular}{ c c }

      \toprule
      \textbf{Method} & \textbf{mIoU(\%)} \\
      \toprule
     pixel-to-pixel&74.5\\
     pixel-to-NROI&\textbf{75.0}\\
      \toprule
    \end{tabular}

  \end{minipage}

    \vspace{-10pt}
  \end{table}
\begin{table}
\small
      \caption{Ablations of the clustering (\textbf{Clusters}) and memory bank (\textbf{Memo.}). \textbf{Left:} We compare mIoUs on Pascal Voc train set when setting $K$ (the number of clusters) with different values. \textbf{Right:} we evaluate the effect of memory bank size.}
  \label{tab:ablation_cluster_memo}
\centering
  \begin{minipage}[b]{4cm}
    \centering
    \begin{tabular}{ c c }
      \toprule
      \textbf{Clusters} & \textbf{mIoU (\%)} \\
      \midrule
      K=4 & 74.6\\
      K=6 & 74.8\\
      K=8 & \textbf{75.0}\\
      K=16& 74.6\\
      \bottomrule
    \end{tabular}
  \end{minipage}
  \begin{minipage}[b]{4cm}
    \centering
    \begin{tabular}{ c c }
      \toprule
      \textbf{Memo.} & \textbf{mIoU (\%)} \\
      \midrule
      $3\times 10^{4}$&74.5\\
      $5\times 10^{4}$&\textbf{75.0}\\
     $8\times 10^{4}$&74.9\\
     $10^{5}$&74.6\\
      \bottomrule
    \end{tabular}

  \end{minipage}

  \vspace{-15pt}
\end{table}

\begin{figure}[t]
\begin{center}
\includegraphics[width=8.3cm]{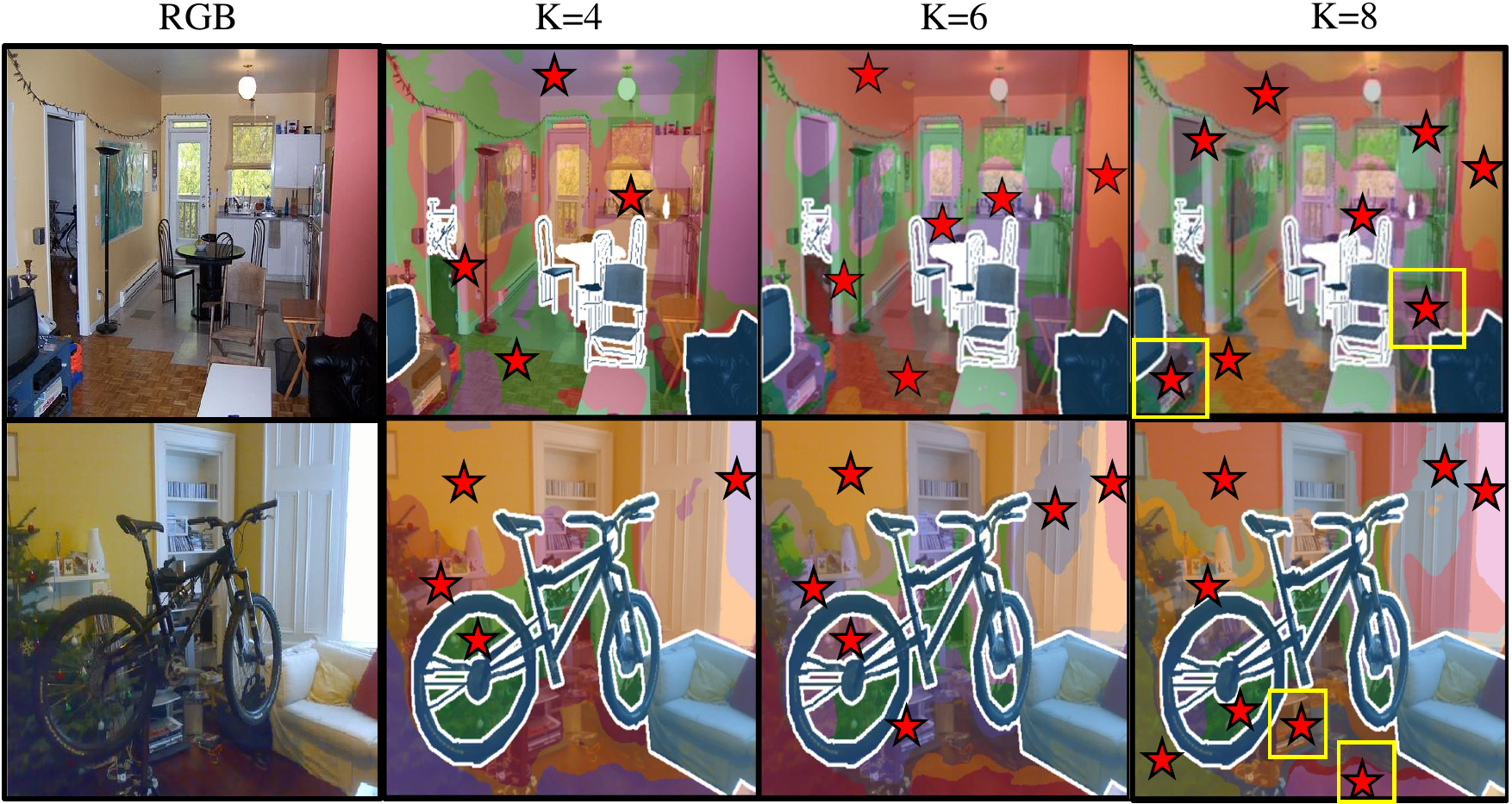}
\end{center}
 \vspace{-10pt}
   \caption{Example results of cluster maps and NROIs. We visualize the clusters and NROIs in different cases. \textbf{From left to right:} RGB images and cluster maps (when $K$=4/6/8). In each cluster, we use the pixel closest (measured with the square distance computed along the feature channel) to the centroid as NROI (marked with the star); the white cropped area is the foreground. The yellow boxes highlight the semantics that are newly recognized with the increasing $K$.
   }
   \label{fig:NROI_visualization}
    \vspace{-10pt}
\end{figure}
\begin{figure}[t]
\begin{center}
\includegraphics[width=8.3cm]{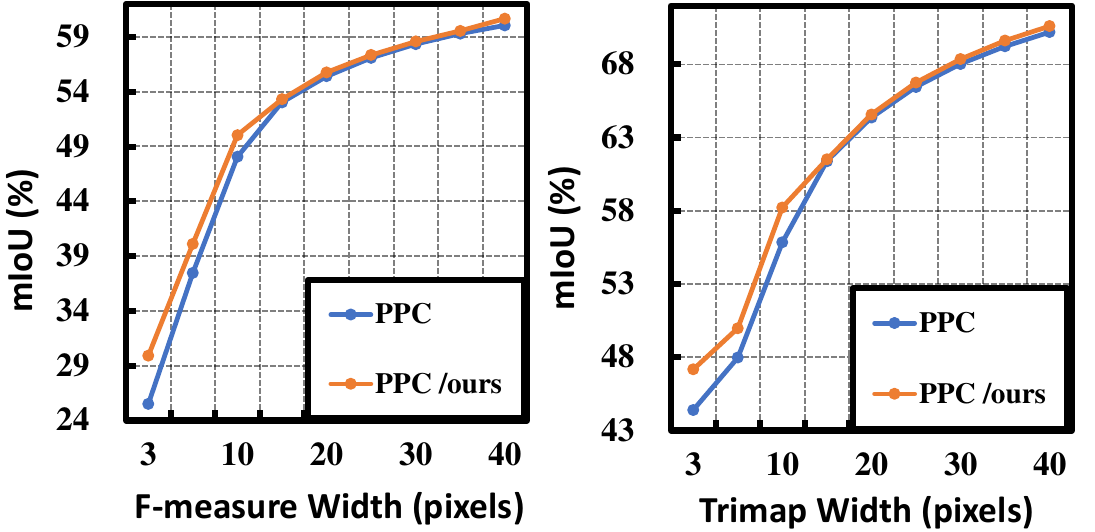}
\end{center}
 \vspace{-10pt}
   \caption{Contour evaluation of seeds on Pascal Voc 2012 train set. The figures show how F-measure and Trimap mIoU results vary with the pixel width of the ground-truth boundary region.
   }
   \label{fig:boundary_evaluation}
    \vspace{-10pt}
\end{figure}
\subsection{Ablation Study} 
\label{sec:ablation}
Below we test the effectiveness of each component and design. All results are the averages over five runs.\\
\textbf{Overall ablation result.} In \refTab{tab:abl_fg_bg}, we ablate the FB and the IF contrastive learning in sequence. For the baseline PPC~\cite{wseg} that estimates a general background prototype to proceed with contrastive learning, the 2nd row (excluding background) shows 0.3\% mIoU gap (73.3\%$\rightarrow$73.6\%). This confirms our assumption that a single prototype is incapable of describing the image background and may even negatively influence foreground classes. The proposed FB and IF contrastive learning contribute 1.4\% mIoU (73.3\%$\rightarrow$74.7\%) and 0.6\% mIoU improvements (73.3\%$\rightarrow$73.9\%). Besides, the auxiliary segmentation loss $L_{seg}$ facilitates better FB and brings an extra 0.3\% mIoU gain (74.7\%$\rightarrow$75.0\%). Overall, our FBR approach improves the baseline by 2.2\% mIoU. \\
\noindent\textbf{Effect of NROI.} We ablate the design and method, and analyze the improvement source to assess the effect:
\begin{itemize}
    \item Projection head $\varphi_{bg}$:  Unlike existing studies implementing contrastive learning~\cite{u2pl,SimCLR} in a common space, we consider the semantic discrepancy between the image foreground and background, i.e., the background has finer semantic granularities. We argue that one representation space is insufficient to generate two different primitives, i.e., NROI and foreground prototype. $\varphi_{bg}$ maps the background semantics to an independent space from the foreground classes, enabling expressive projection and better semantic representation. In \refTab{tab:ablation_NROI} (Left), we compare the single-head design, namely using one projection head $\varphi_{fg}$ to represent both foreground and background, with ours; the 0.7\% mIoU gain (74.0\% $\rightarrow$74.7\%) verifies the benefit of our design.

    \item Contrast strategy: We compare our pixel-to-NROI contrast with the brute-force strategy (\refFig{fig:NROI_demonstration} (a)) that puts all background features of $Z_{bg}$ into the memory bank to compute $L_{pcl}^{bg}$, a.k.a. pixel-to-pixel. To make a fair comparison, we set the bank size to 100$\mathbf{K}$ (causing longer sampling time) in the brute-force case, yet set ours to 50$\mathbf{K}$. In \refTab{tab:ablation_NROI} (Right), ours outperforms the pixel-to-pixel contrast by 0.5\% mIoU, showing NROI's benefits against the common pixel feature-based representation.
    \item Source of performance gain: The improvements obtained through NROIs mainly stem from the fine-grained background semantic recognition. In \refFig{fig:bg_tsne} and \refFig{fig:bg_umap}, we respectively employ t-SNE~\cite{tsne} and UMAP~\cite{umap} for background-foreground feature analysis. Detailly, we compare features of PPC~\cite{wseg} (from its $f_{proj}$) before and after using FB contrast. Although the projected manifolds by these two methods have different shapes due to the difference in dimension reduction technique, we get consistent findings: in (a) of both figures, the scatted background features result in the spurious semantic correlation; foreground features are contaminated with confusing intra- and inter-image background information, e.g., the overlapped part in the boxes. In contrast, (b) adopts NROIs to capture the fine-grained background semantics and begets a structured background feature space. Furthermore, contrastive optimization suppresses the suspicious background cues and distinguishes them from the foreground, effectively avoiding confusion. Additionally, we select three representatives of co-occurring background semantics (sky, lake, and railroad) and compare CAMs in \refFig{fig:co_occurring_refinement}. FB contrast resists background disturbances and obtains more reliable results. 
\end{itemize}

\noindent\textbf{Hyperparameter discussion.} In this part, we discuss the effect of NROI's hyperparameters:
\begin{itemize}
    \item Number of clusters: In \refTab{tab:ablation_cluster_memo} (Left), we ablate the cluster number $K$ and report the seed accuracy. Intuitively, increasing $K$ will return finer-grain background semantics, but make NROIs less meaningful and increase the computational cost. Based on the comparison result, we set $K=8$ throughout the paper. Besides, we visualize the clustering maps and NROIs in \refFig{fig:NROI_visualization}. We see that clustering groups the image background and fine-grained semantics are identified with the increasing $K$. For instance, the “TV cabinet" and the “table" (the 1st row) and the “book" and “floor" (the 2nd row). As for the extracted NROIs, i.e., the cluster centroids, indicating the most typical representation of the individual semantic group, usually located in the object-central region, capturing meaningful background semantics, with which we can model background content explicitly.
    \item Size of memory bank: In \refTab{tab:ablation_cluster_memo} (Right), we investigate the effect of the memory bank size. The larger bank can store more NROIs yet also indicates a longer sampling time. Therefore, we set the bank size to 50$\mathbf{K}$, 200$\mathbf{K}$ for Pascal Voc 2012 and MS COCO 2014.
    \item Loss weights: In \refTab{tab:loss_weight}, we conduct ablation studies on the loss weights $\lambda_1$ and $\lambda_2$ to evaluate the effects of the two components of the contrastive loss $L_{pcl}$ (in Eq. \ref{eq:full_loss_formulation}):  $L^{bg}_{pcl}$ and $L^{fg}_{pcl}$. A higher value of $\lambda_1$ suggests that $L_{pcl}$ would focus more on the fore-to-background relationship, and vice versa. Our results indicate that FBR performs best when $\lambda_{1}$ and $\lambda_{2}$ reach 0.10 and 0.01, respectively. Besides, the comparison results exhibit that our FBR method is more sensitive to $\lambda_{1}$ than $\lambda_{2}$, showing the importance of fore-to-background contrastive learning.

\end{itemize}
\begin{table}
    \centering
    \small
        \caption{Effects of loss weight $\lambda_{1}$ and $\lambda_{2}$ on mIoU (\%) results of the seeds, seeds w/ CRF, and pseudo masks (\textbf{Mask}) on Pascal Voc 2012 train set. $\lambda_{1}$ and $\lambda_{2}$ are used to balance the $L_{pcl}^{bg}$ and  $L_{pcl}^{fg}$ in $L_{pcl}$ (Eq. \ref{eq:full_loss_formulation}), respectively.}
    \begin{tabu}{ccccc}
   \tabucline[1pt]{-} 
    \textbf{$\lambda_{1}$}&\textbf{$\lambda_{2}$}& \textbf{Seed}&\textbf{+CRF}&\textbf{Mask}\\
    \tabucline[1pt]{-} 
    \multirow{3}{*}{0.05}&0.01&71.3&74.0&74.7\\
    ~&0.05&71.6&74.8&75.1\\
    ~&0.10&72.1&74.8&75.4\\
    \hline
        \multirow{3}{*}{0.10}&0.01&\textbf{72.2}&\textbf{75.5}&\textbf{75.9}\\
        
    ~&0.05&71.7&74.7&75.3\\
    ~&0.10&71.7&74.7&75.2\\
    \hline
        \multirow{3}{*}{0.20}&0.01&71.5&74.2&74.9\\
    ~&0.05&70.9&73.7&74.4\\
    ~&0.10&70.8&73.5&74.1\\
    \tabucline[1pt]{-} 
    \end{tabu}
    \label{tab:loss_weight}
    \vspace{-10pt}
\end{table}

\noindent\textbf{Effect of active sampling.} In this part, we evaluate our active method and analyze its effectiveness:
\begin{table}
\centering
      \caption{Negative sampling method comparison~\cite{exploring_hard_sampling,wseg} (mIoU (\%)) in terms of seeds (\textbf{Seed}) and seeds with CRF (\textbf{+CRF}) on Pascal Voc 2012 train set.}
   \label{tab:ablation_active_sampling}
\small
    \begin{tabu}[t]{m{4.4cm}<\centering m{1.2cm}<\centering
    m{1.2cm}<\centering}
    \tabucline[1pt]{-} 
      \textbf{Method} & \textbf{Seed}&\textbf{+CRF} \\
    \tabucline[1pt]{-}
      PPC (hard, original) & 70.5&73.3\\
            PPC (active, ours) & \textbf{71.1\tiny{+0.6}}&\textbf{73.9\tiny{+0.6}}\\
    \tabucline[1pt]{-} 
    \end{tabu}

\end{table}

\begin{table}
\centering
      \caption{Evaluate the sampling methods under different query settings. The hard sampling experiments are implemented with PPC~\cite{wseg}. \textbf{naive query:} all pixels are queries.
      \textbf{hard query:} half hard queries and half random queries \cite{wseg,exploring_hard_sampling}.
      \textbf{adaptive query:} select queries based on Eq. \ref{eq:naive_query}.}
   \label{tab:appendix_active_sampling}
\small
    \begin{tabu}[t]{m{4.4cm}<\centering m{1.2cm}<\centering
    m{1.2cm}<\centering}
    \tabucline[1pt]{-} 
      \textbf{Method} & \textbf{Seed}&\textbf{+CRF} \\
    \tabucline[1pt]{-}
      baseline (EPS \cite{eps}) & 69.5&71.4\\
      +hard sampling (naive query) & 70.4&73.2\\
      +hard sampling (hard query) & 70.5&73.3\\
      +hard sampling (adaptive query) & \textbf{70.5\tiny{+(1.0)}}&\textbf{73.4\tiny{+(2.0)}}\\
      +active sampling (naive query)& 70.8&73.5\\
        +active sampling (hard query)& 71.0&73.7\\
            +active sampling (adaptive query)& \textbf{71.1\tiny{+(1.6)}}&\textbf{73.9\tiny{+(2.5)}}\\
    \tabucline[1pt]{-} 
    
    \end{tabu}
 \vspace{-10pt}
\end{table}

\begin{itemize}
    \item Active sampling $vs.$ hard sampling: We compare the active method against the hard sampling~\cite{exploring_hard_sampling,wseg} that selects negative keys based on the “hardness" computation. In \refTab{tab:ablation_active_sampling}, after replacing the hard sampling (adopted in PPC) with ours, the baseline is improved by 0.6\% mIoU. Besides, we additionally experiment with both sampling methods on EPS and report extensive comparison results (in \refTab{tab:appendix_active_sampling}) under different query settings: naive query, hard query, and adaptive query.  We observe that the active method consistently improves the baseline and always outperforms the hard sampling method, showing its robustness. Particularly, when proceeding with IF contrast (i.e., $L_{pcl}^{fg}$) with the adaptive query setting (in Eq. \ref{eq:naive_query}), our method improves EPS by 1.6\% mIoU on seed and 2.5\% mIoU on masks (against the maximum 1.0\%/2.0\% mIoU $\uparrow$ achieved by the hard sampling method~\cite{wseg}). More importantly, our active sampling draws the negatives on the fly, without the hardness calculation and the sorting, resulting in a lower computation cost. 
    \item Activate complete object regions: Considering that background is present in nearly every image and the inaccuracies in seed results, our active approach excludes the background class during negative sampling. This allows us to concentrate on optimizing relationships within the foreground and thus learn more discriminative class features. When combined with IF contrast,  we prevent interference from suspicious background pixel features (false positives) in the sampling, particularly those in the transition area between the foreground and background. This way, we learn more accurate object contours. In \refFig{fig:fg_tsne}, we visualize the foreground  feature space via t-SNE~\cite{tsne} and UMAP~\cite{umap}. After the sampling method replacement, the cropped class features become more compacted, e.g., “car" and “chair". This observation presents our IF contrast's effectiveness in learning discriminative FG features. Besides, we employ F-measure~\cite{F_measure} and Trimap~\cite{DeepLabv2, EPL} to evaluate the object contour quality of the generated seeds. Given a pixel width, both metrics assess the alignment degree between the prediction and its ground truth within a narrow band region from the true semantic boundary. In \refFig{fig:boundary_evaluation}, our obtained performance gain on PPC~\cite{wseg} mainly in the region near the object contour (pixel width $\leq10$), indicating our IF contrast's effects in learning complete object regions. 
\end{itemize}

Compared with existing approaches~\cite{wseg,reco}, FBR does not require cross-view inputs and incurs no inference overhead. Our results (\refTab{tab:abl_fg_bg} \& \refFig{fig:bg_tsne}) verify that the major improvements come from the proposed NROI-based FB contrastive learning, significantly improving CAMs' ability in background discrimination and thus avoiding classification ambiguity. Our FBR has high versatility (across different baselines) thanks to its simplicity in design and modularized techniques; the advanced segmentation results manifest FBR's usefulness.

\begin{figure}
    \centering
    \includegraphics[width=\linewidth]{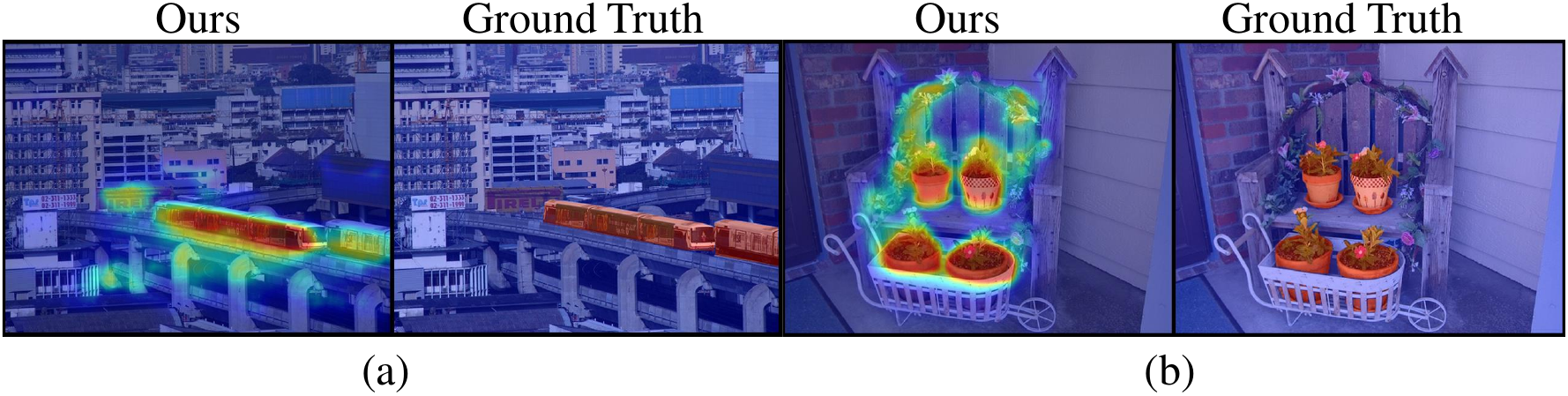}
     \vspace{-10pt}
    \caption{Failure cases. We provide CAMs of (a) “train” and  (b) “potted plants” and compare our FBR with the ground truth.}
    \label{fig:failure_analysis}
    \vspace{-15pt}
\end{figure}

\noindent\textbf{Limitations.} Despite achieving SOTA results in various WSSS tasks, our approach encounters challenges in two cases: 1) when the image background is visually too similar to the foreground classes and 2) when the foreground classes have irregular object contours.  As shown in \refFig{fig:failure_analysis}, the background buildings are misclassified as a part of “train” in (a) upon confusing appearances. Besides, our method can not accurately predict the boundary of “potted plants” (b) due to their complex shapes. A plausible explanation for these failures is that our method focuses on correcting low-confidence regions in CAMs, while these challenging regions already exhibit high activation scores during the classification process and are thus ignored. This problem can be mitigated with a powerful backbone, e.g., ViT, to obtain more precise seed predictions.

\noindent\textbf{Future works.} An intriguing direction is to integrate our approach with foundation models like SAM~\cite{SAM}. Leveraging SAM's object mask predictions, our FBR could accurately label the background/foreground semantics, generating pseudo-labels with high-quality object contours. These pseudo-labels, enriched with precise boundary information, could benefit segmentation practices in other domains, such as medical or remote sense images. However, an adaptive clustering method would be required to accurately determine the number of background semantics, which we leave for future development.

%% file: Content/Conclusion.tex
This paper proposed a simple fine-grained background representation method, FBR, to address the co-occurring background problem and learn integral object masks in weakly supervised semantic segmentation (WSSS). Our method designs a new background primitive and an active sampling method to perform the fore-to-background and intra-foreground contrastive learning. Extensive experiments on Pascal Voc and MS COCO demonstrated the good merits of FBR in generating pseudo masks, achieving new state-of-art performances in WSSS, and also benefiting the instance-level segmentation.

%% file: Content/Appendix.tex
This appendix reports the details of the experiment setting and implementation and provides additional ablation studies and qualitative results.

\begin{figure}[h]
    \centering
    \includegraphics[width=8.3cm]{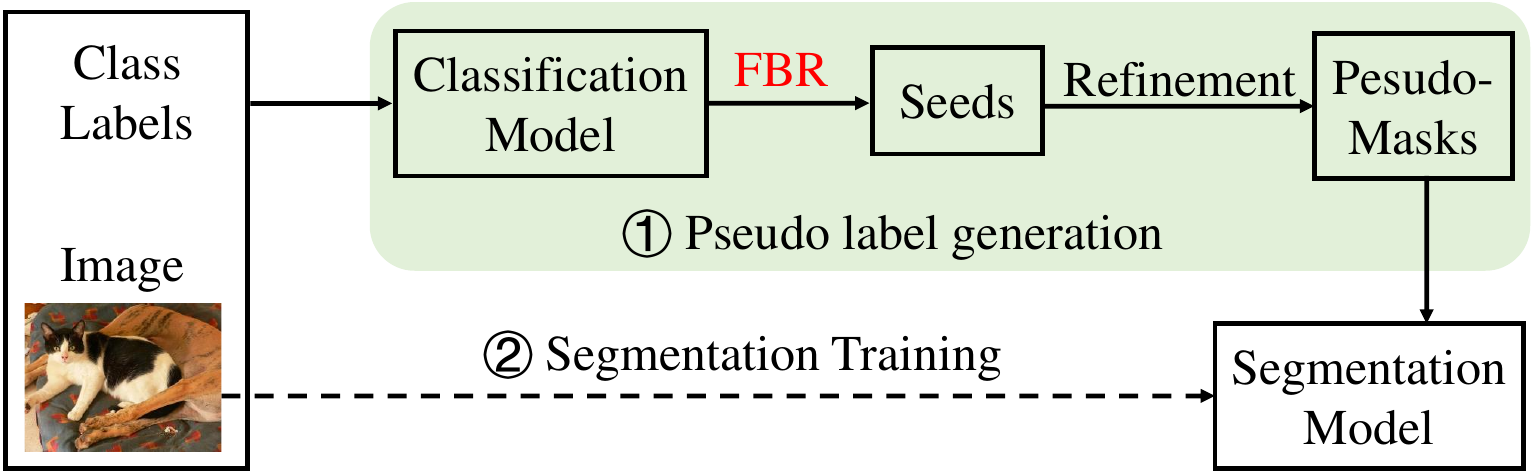}
    \caption{Overall training pipeline. Our contribution (FBR method) is to improve the classification model to enhance CAMs and generate more precise seed predictions.}
    \label{fig:train_pipeline}
    \vspace{-10pt}
\end{figure}

\subsection{Experiment details}

\noindent\textbf{Overall training pipeline.} As shown in \refFig{fig:train_pipeline}, we first employ our FBR method on the classification model to enhance the seed predictions by addressing the co-occurring background problem and learning integral object regions. After refining [1] the seeds to obtain the pseudo masks, we utilize them to train the segmentation model and get the final results.

\noindent\textbf{Seed generation.} We follow the original configurations reported in PPC~\cite{wseg}, SEAM~\cite{seam}, and AMN~\cite{AMN} for our baseline experiments. Based on the ablation results in ~\refTab{tab:additional_ablations}, we set the loss weights of the background pixel-wise cross-entropy loss ($L_{pcl}^{bg}$), foreground pixel-wise cross-entropy loss ($L_{pcl}^{fg}$), and segmentation loss ($L_{seg}$) to 0.1, 0.01, and 0.01, respectively, for PPC and SEAM. For AMN, we set these loss weights to 0.1, 0.05, and 0.01.
When applying our FBR to PPC and SEAM, we employ a learning rate (lr) of 0.01 and a batch size of 9, optimized using the PolyOptimizer. For experiments on AMN, we adopt a lr of 5e-6, a batch size of 1, and utilize the Adam optimizer.
Furthermore, when experimenting on the MS COCO and Pascal VOC datasets, we set the feature bank size to 200 K and 50 K, respectively.

\noindent\textbf{Semantic segmentation.} For the segmentation practice, we strictly follow the settings of the baselines. Training images are randomly scaled in the range of [0.5, 0.75, 1.0, 1.25, 1.5] and cropped to 321$\times$ 321 for Pascal Voc 2012, 481$\times$ 481 for MS COCO 2014. We adopted the SGD optimizer (lr=0.01) and set the batch size to 10 (16 for MS COCO 2014). The number of training steps is 30$K$ (100$K$ for COCO).\\
\noindent\textbf{Instance segmentation.} We follow the training settings of BoxInst [38], except resizing the shorter side of images in [480,640]. In training, we set the batch size to 16 and use the SGD optimizer (with learning rate = 0.01).
\subsection{Additional ablation analysis}

\noindent\textbf{Computational overhead \& Complexity.} Utilizing PPC~\cite{wseg} as the baseline model, we conduct a comprehensive analysis to evaluate the computational efficiency of our FBR approach.

Based on PPC, FBR slightly increases the training hours (4.8 h $\rightarrow$ 6.4 h) and model size (102.9 M $\rightarrow$ 103.9 M). The increased computation cost comes from the added background projection head $\varphi_{bg}$ and the classifiers $\varphi_{seg}$. Notably, our method does not affect the inference speed of the baseline since all involved modules are removed at the test time.

\begin{table}[h]
\small
    \caption{We compare the computational cost of PPC~\cite{wseg} and our method regarding the parameter size (million, M), training time (hours, h), inference speed (second / per image), and GPU memory footprint (GB).}
    \label{tab:comparing_inference_cost}
    \centering
    \begin{tabu}{c|cccc}
   \tabucline[1pt]{-} 
         Method&Param. &Train. time&Infer. speed & Memory\\
  \tabucline[1pt]{-} 
         PPC [10]&102.9 M& 4.8 h & 1.86 s &14.9 GB\\
         \rowcolor{Gray}Ours&103.9 M& 6.4 h&1.86 s &16.5 GB\\ 
\tabucline[1pt]{-} 
    \end{tabu}
    \vspace{-10pt}
\end{table}

\noindent\textbf{Auxiliary BG segmentation.} To obtain effective background negative samples, we define a learning objective for $Z_{bg}$, i.e., the background segmentation, requiring $Z_{bg}$ to discriminate the background region and learn more representative features.

We consider pixels with a summed CAM score (on foreground classes) smaller than 0.05 as the background to conduct the auxiliary segmentation training. We add a loss weight $\alpha$ before the background segmentation loss $L_{seg}$ in Eq. 10, and ablate $\alpha$ (test on Pascal Voc 2012 train set) in \refTab{tab:additional_ablations}. We observe that our method performs best when setting $\alpha=0.01$. 

\begin{table}[h]
\small
      \caption{Additional ablation experiments of the background segmentation. $\alpha$ is the loss weight of $L_{seg}$ in Eq. 10. }
    \centering
    \begin{tabu}{ c| ccc c }
    \tabucline[1pt]{-} 
      \textbf{$\alpha$} & 0 &0.01 & 0.025&0.05 \\
      \hline
      mIoU&75.2&\textbf{75.5}&74.9&74.1\\
 \tabucline[1pt]{-} 
    \end{tabu}

\vspace{-15pt}

  \label{tab:additional_ablations}
\end{table}
\noindent\textbf{Effects of TAP.} Proposed in ~\cite{tap}, TAP only aggregates above-threshold pixels in the semantic feature $f$ to compute the classification score. We express TAP as follows:
\begin{equation}
    s^{tap}_{c}=\sum_{i=1}^{L}\theta_{c,i}\frac{\sum_{j\in\Omega}\mathbbm{1}[f_{i,j}>\alpha] f_{i,j}}{\sum_{j\in\Omega}\mathbbm{1}[f_{i,j}>\alpha]},
    \label{eq:tap}
\end{equation}
where $\alpha$ is the threshold (set to 0.1) and $\Omega$ is the coordinate set of $\mathbb{R}^{H\times W}$. TAP has been proven to be better at filtering out over-activated BG regions [4] than GAP, thus generating more accurate seeds. In our work, TAP brings more accuracy improvements than GAP (74.7\% $v.s$ 75.0\%) when assembling our FBR method on PPC [10].

\subsection{Additional Experimental Results}
\begin{table*}[t]
\scriptsize
\centering
\caption{(\textbf{Pseudo Masks}) Category-level mIoU comparison on Pascal Voc 2012 train set. Highlights are classes in which we achieve up to 2.5\% mIoU improvements.}
\begin{tabu}{m{0.7cm}<{\centering}|m{0.5cm}<{\centering}|m{0.3cm}<{\centering}m{0.3cm}<{\centering}m{0.3cm}<{\centering}m{0.3cm}<{\centering}m{0.3cm}<{\centering}m{0.3cm}<{\centering}m{0.3cm}<{\centering}m{0.3cm}<{\centering}m{0.3cm}<{\centering}m{0.3cm}<{\centering}m{0.3cm}<{\centering}m{0.3cm}<{\centering}m{0.3cm}<{\centering}m{0.3cm}<{\centering}m{0.3cm}<{\centering}m{0.3cm}<{\centering}m{0.3cm}<{\centering}m{0.3cm}<{\centering}m{0.3cm}<{\centering}m{0.3cm}<{\centering}m{0.3cm}<{\centering}}
    \tabucline[1pt]{-} 
\textbf{Method} &\textbf{mIoU}&\textbf{bgr}&\textbf{aero}&\textbf{bicy}&\textbf{bird}&\textbf{boat}& \textbf{bottle}&\textbf{bus}&\textbf{car}&\textbf{cat}&\textbf{chair}&\textbf{cow}&\textbf{dt.}&\textbf{dog}&\textbf{horse}&\textbf{motor}& \textbf{pers.}&\textbf{pott}&\textbf{sheep}&\textbf{sofa}&\textbf{train}&\textbf{tv.}\\
\tabucline[1pt]{-}  
AMN& 72.2&90.2&75.3&40.1&77.4&67.9& 73.4&85.6&78.9&80.7&36.5&86.1&65.8&78.7&83.4&81.0&74.4&62.4&89.4&62.8&65.3&63.1\\
+ours&\textbf{73.1}&90.8&74.4&\textbf{45.1}&\textbf{84.9}&69.9&71.7&84.4&79.3&\textbf{87.1}&37.2&85.6&61.6&\textbf{84.5}&81.0&79.2&73.8&63.9&89.8&63.5&66.3&60.2\\
\hline
PPC& 73.3&91.2&86.6&44.6&82.8&80.9&73.1&84.0&81.4&88.9& 31.2&83.7&52.7&85.6&86.9&81.7&80.5&54.2&85.9&52.5&77.6&53.8\\
+ours&\textbf{75.9}&92.4&86.4&\textbf{47.7}&\textbf{85.0}&83.1&75.1&85.0&\textbf{85.5}&\textbf{91.6}&39.7&\textbf{88.3}&50.6&\textbf{91.6}&\textbf{90.7}&83.6&81.1&\textbf{63.2}&\textbf{90.0}&48.5&\textbf{83.9}&51.7\\
\tabucline[1pt]{-} 
\end{tabu}

\label{tb:perclass_pascal_masks}
\end{table*}
\begin{table*}[t]
\centering
\scriptsize
 \vspace{-10pt}
\caption{(\textbf{Semantic Segmentation}) Category-level mIoU comparison on Pascal Voc 2012 test set. Highlights are classes in which we achieve up to 3.0\% mIoU improvements. The reported AMN performance is from the ImageNet pre-trained model. }
\begin{tabu}{m{0.8cm}<{\centering}|m{0.5cm}<{\centering}|m{0.3cm}<{\centering}m{0.3cm}<{\centering}m{0.3cm}<{\centering}m{0.3cm}<{\centering}m{0.3cm}<{\centering}m{0.3cm}<{\centering}m{0.3cm}<{\centering}m{0.3cm}<{\centering}m{0.3cm}<{\centering}m{0.3cm}<{\centering}m{0.3cm}<{\centering}m{0.3cm}<{\centering}m{0.3cm}<{\centering}m{0.3cm}<{\centering}m{0.3cm}<{\centering}m{0.3cm}<{\centering}m{0.3cm}<{\centering}m{0.3cm}<{\centering}m{0.3cm}<{\centering}m{0.3cm}<{\centering}m{0.3cm}<{\centering}}
    \tabucline[1pt]{-} 
\textbf{Method} &\textbf{mIoU}&\textbf{bgr}&\textbf{aero}&\textbf{bicy}&\textbf{bird}&\textbf{boat}& \textbf{bottle}&\textbf{bus}&\textbf{car}&\textbf{cat}&\textbf{chair}&\textbf{cow}&\textbf{dt.}&\textbf{dog}&\textbf{horse}&\textbf{motor}& \textbf{pers.}&\textbf{pott}&\textbf{sheep}&\textbf{sofa}&\textbf{train}&\textbf{tv.}\\
\tabucline[1pt]{-} 

AMN& 69.6&90.7&82.8&32.4&84.8&59.4& 70.0&86.7&83.0&86.9&30.1&79.2&56.6&83.0&81.9&78.3&72.7&52.9&81.4&59.8&53.1&56.4\\
+ours&\textbf{73.2}&91.3&82.9&\textbf{35.3}&\textbf{90.5}&59.1&70.1&\textbf{90.1}&84.0&\textbf{91.2}&\textbf{36.5}&\textbf{85.9}&\textbf{66.3}&\textbf{88.8}&\textbf{87.3}&79.1&\textbf{77.2}&\textbf{63.6}&\textbf{86.1}&59.7&53.4&58.4\\
\hline
PPC& 73.6&92.1&92.3&40.6&89.8&65.4&69.9&91.5&83.6&90.9& 31.4&86.2&48.2&85.1&89.8&81.9&80.2&59.6&87.7&52.9&80.3&46.4\\
+ours&\textbf{74.9}&92.2&92.6&40.7&86.4&63.6&70.1&92.1& 84.2&91.3&\textbf{36.1}&\textbf{88.2}&\textbf{51.6}&\textbf{89.3}&89.9&83.2&78.1&\textbf{72.7}&90.1&55.1&79.3&45.3\\
\tabucline[1pt]{-} 
\end{tabu}
\vspace{-10pt}

\label{tb:perclass_pascal_segmentation}
\end{table*}
\noindent\textbf{Category-level Evaluation}
\refTab{tb:perclass_pascal_masks} shows category-level mIoU evaluation in terms of the pseudo masks. We observe that FBR significantly enhances baseline models' ability to distinguish classes that have complicated shapes, like “bird", “dog", and “dining table (dt.)." Also, we report the segmentation results in \refTab{tb:perclass_pascal_segmentation} after training the segmentation network with the pseudo masks and observe considerable performance improvements on the baselines.

\begin{figure}[h]
\begin{center}
\includegraphics[width=8.3cm]{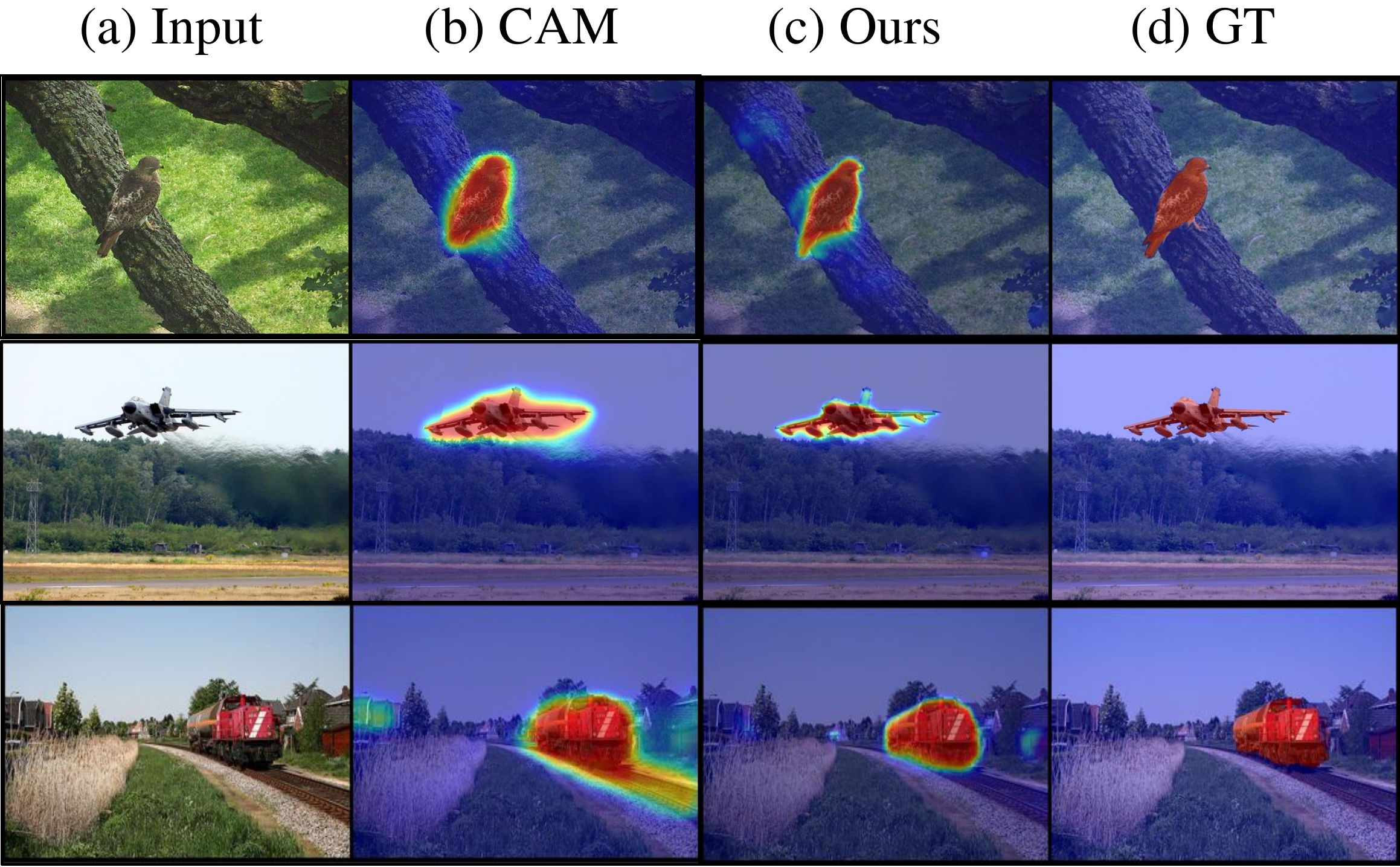}
\end{center}
 \vspace{-10pt}
   \caption{Example CAM results (solve co-occurring BG). Left to Right: (a) input image, (b) CAM results generated by AMN, (c) CAM results from AMN W/ ours, and (d) the ground truth.
   }
   \label{fig:appendix_results1}
   \vspace{-10pt}
\end{figure}

\begin{figure}[h]
\begin{center}
\includegraphics[width=8.3cm]{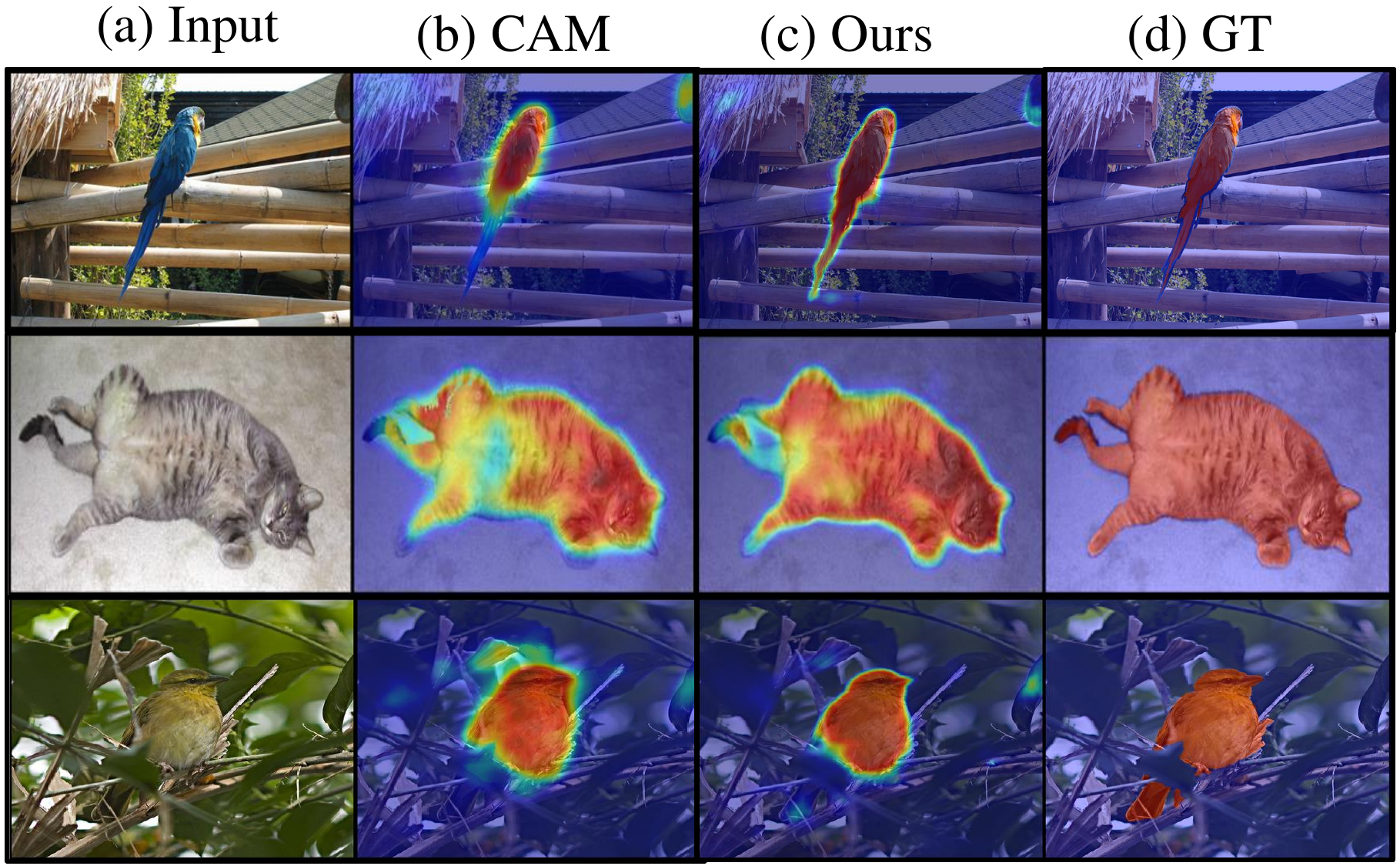}
\end{center}
 \vspace{-10pt}
   \caption{Example results (activating more object regions).
   }
   \label{fig:appendix_results2}
      \vspace{-15pt}
\end{figure}

\begin{figure}[h]
\begin{center}
\includegraphics[width=8.3cm]{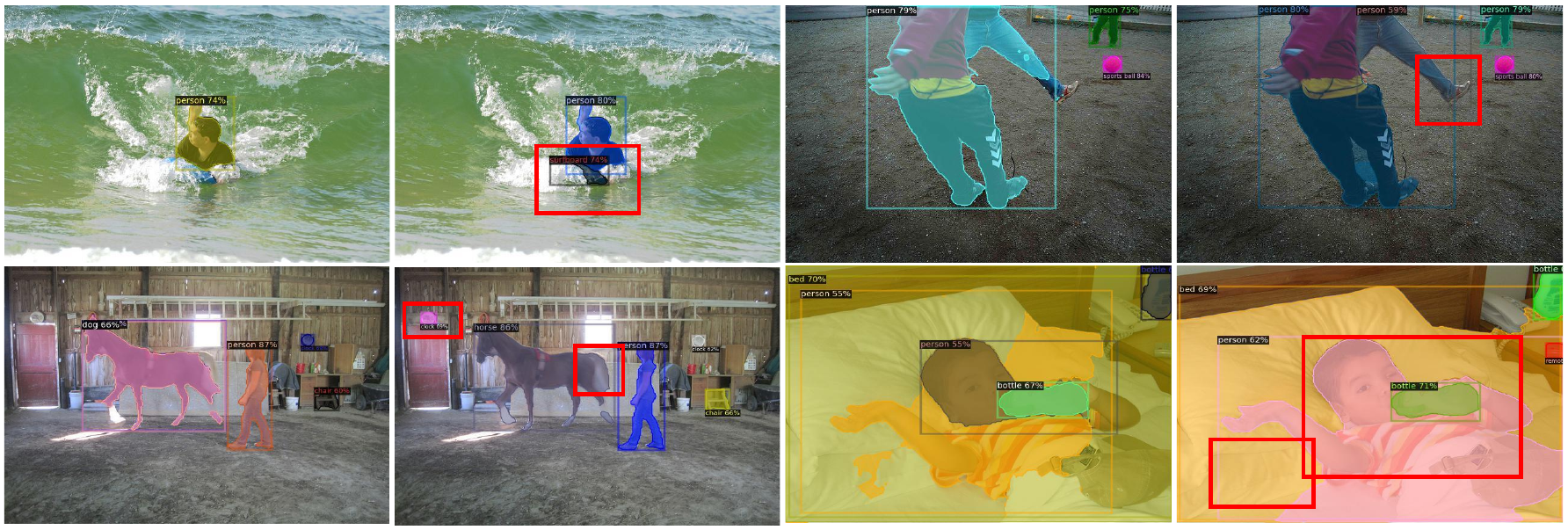}
\end{center}
 \vspace{-10pt}
   \caption{Example results of instance segmentation.
   }
   \label{fig:appendix_results2}
      \vspace{-15pt}
\end{figure}

\noindent\textbf{Quantitative results} \refFig{fig:appendix_results1} and \refFig{fig:appendix_results2} provide more CAM results to show FBR's effectiveness. In \refFig{fig:appendix_results1}, we see that FBR effectively helps the baselines distinguish the target FG object from the confusing BG semantics. In~\refFig{fig:appendix_results2}, we notice that the baseline (with FBR) learns more class features, thus activating more integral object regions.